\documentclass[10pt]{article}

\usepackage[
  letterpaper,
  textwidth=5.5in,
  textheight=9in,
  top=1in,
  headheight=30pt,
  headsep=12pt,
  footskip=30pt
]{geometry}
\usepackage{natbib}

\usepackage[utf8]{inputenc}
\usepackage[T1]{fontenc}
\usepackage{amsfonts}
\usepackage{amsmath}
\usepackage{amssymb}
\usepackage{booktabs}
\usepackage{graphicx}
\usepackage{float}
\usepackage{fancyhdr}
\usepackage{longtable}
\usepackage{microtype}
\usepackage{subcaption}
\usepackage{xcolor}
\usepackage{url}
\usepackage{hyperref}
\hypersetup{
  hidelinks,
  pdftitle={LM-X: Explainable Vision-Language-Action Modeling via Progress, Event, and Uncertainty Prediction},
  pdfauthor={Jin Lou, Zhiyuan Jing, Xupeng Wang, Andong Chen, Xingdong Zhu, Yuexuan Li, Yuan Xu, Zhijie Zhu, Yingwei Ji, Wenpeng Nie, Renxing Feng, Liangliang Chen, Ying Chu, Jingyi Li, Jinyan Liu, Zhiqi Song, Jingxuan Zhu, Jidong Zhang, Yufei Liu, Boyang Xing, Lei Jiang, Yan Cui, Hongming Li, Yuchen Zhu}
}
\setcitestyle{authoryear,round,semicolon}

\newcommand{\method}{\textsc{LM-X}}

\newcommand{\affillogo}[1]{\raisebox{-0.23\height}{\includegraphics[height=3.1mm]{#1}}}

\fancypagestyle{firstpage}{%
  \fancyhf{}
  \fancyhead[L]{\includegraphics[height=8.5mm]{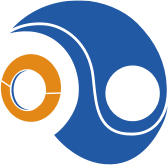}}
  \fancyhead[R]{\includegraphics[height=8.5mm]{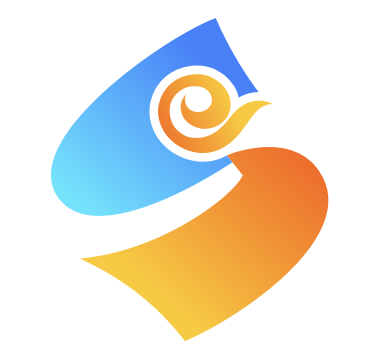}}
  \fancyfoot[C]{\thepage}

}

\title{LM-X: Explainable Vision--Language--Action Modeling via Progress, Event, and Uncertainty Prediction}

\author{Jin Lou, Zhiyuan Jing, Xupeng Wang, Andong Chen, Xingdong Zhu, Yuexuan Li, Yuan Xu, Zhijie Zhu, Yingwei Ji, Wenpeng Nie, Renxing Feng, Liangliang Chen, Ying Chu, Jingyi Li, Jinyan Liu, Zhiqi Song, Jingxuan Zhu, Jidong Zhang, Yufei Liu, Boyang Xing, Lei Jiang, Yan Cui, Hongming Li, Yuchen Zhu}
\date{}

\begin{document}

\thispagestyle{firstpage}
\begin{center}
  \vspace*{-0.5em}
  {\fontsize{16.5}{19.5}\selectfont\bfseries LM-X: Explainable Vision--Language--Action Modeling\\
  via Progress, Event, and Uncertainty Prediction\par}
  \vspace{0.85em}
  {\small
  Jin Lou$^{1,*}$, Zhiyuan Jing$^{2,*}$, Xupeng Wang$^{1}$, Andong Chen$^{1}$\\[-0.05em]
  Xingdong Zhu$^{1}$, Yuexuan Li$^{1}$, Yuan Xu$^{1}$, Zhijie Zhu$^{1}$, Yingwei Ji$^{1}$\\[-0.05em]
  Wenpeng Nie$^{1}$, Renxing Feng$^{1}$, Liangliang Chen$^{2}$, Ying Chu$^{2}$, Jingyi Li$^{2}$\\[-0.05em]
  Jinyan Liu$^{2}$, Zhiqi Song$^{2}$, Jingxuan Zhu$^{2}$, Jidong Zhang$^{2}$\\[-0.05em]
  Yufei Liu$^{1}$, Boyang Xing$^{1}$, Lei Jiang$^{1}$, Yan Cui$^{1}$\\[-0.05em]
  Hongming Li$^{2,*,\dagger}$, Yuchen Zhu$^{1,\dagger}$\par}
  \vspace{0.55em}
  {\footnotesize
  $^{1}$\affillogo{logo_1.png}\; Humanoid Robot (Shanghai) Co., Ltd.\\[0.15em]
  $^{2}$\affillogo{logo_2.png}\; E-surfing Digital Life Technology Co., Ltd., China Telecom\par}
  \vspace{0.35em}
  {\scriptsize
  $^{*}$ Equal contribution.\quad $^{\dagger}$ Corresponding authors.\\[-0.05em]
  \href{mailto:hongmingli1995@gmail.com}{hongmingli1995@gmail.com}\quad
  \href{mailto:zhuyuchen@openloong.net}{zhuyuchen@openloong.net}\par}
\end{center}
\vspace{0.25em}

\begin{abstract}
Large-scale vision--language--action (VLA) policies have advanced generalist robot control, yet most remain stimulus-to-action black boxes: actions are exposed, but their explanatory state is not. They provide no native account of three explanatory signals: task progress, the next semantic transition, or local command reliability. Prior work shows that progress and event structure aid long-horizon control and that uncertainty supports monitoring; however, such capabilities are typically added or extracted only after action pretraining. The field therefore lacks a VLA foundation model whose explanatory state is jointly pretrained with control. Drawing on biological sensorimotor organization, in which outcome-sensitive, event-segmented, and probabilistic predictions structure behavior, we introduce \method{}. \method{} learns three directly supervised online signals: return-to-go (RTG) estimates visible progress and state quality; event-to-go (ETG) predicts the action sequence to the next semantic event; and heteroscedastic action-flow variance reports local command reliability. RTG conditions ETG and both condition action generation; uncertainty is estimated inside the action expert, making explanation part of control rather than a post-hoc description. We pretrain \method{} on more than 20,000 hours of heterogeneous real-robot trajectories, including over 1,000 hours of failed rollouts. A controlled gate favors joint over post-hoc training. \method{} achieves 74.1\% success on 50 randomized-hard RoboTwin2.0 tasks and 73.5\% on seven real-robot tasks, compared with 55.4\% and 50.7\% for GR00T N1.7. Its signals track progress and regression, anticipate event-scale motion, detect high-error actions, and provide advance failure warning. These results establish \method{} as an explainable VLA foundation model that couples transparent predictive state with stronger generalist control. Open-source release can be found at \url{https://github.com/loongOpen/LoongWu-LM-X-VLA}
\end{abstract}

\section{Introduction}
Large-scale VLA policies increasingly transfer visual--language knowledge into closed-loop robot control \citep{brohan2022rt1,brohan2023rt2,openx2023,kim2024openvla,black2024pi0}. Despite their strong performance, most still behave as stimulus-to-action black boxes: observations and instructions enter, and an action chunk emerges, while the policy's assessment of progress, intermediate intention, and local reliability remains implicit. The executed motion is not itself an explanation---the same displacement may be a nominal approach, a correction after a missed grasp, or an oscillation. An operator can observe \emph{what} the robot does, but not whether the policy believes execution is advancing, which meaningful transition it is pursuing, or whether the command is trustworthy \citep{dragan2013legibility,hayes2017transparency,rudin2019interpretable}.

Biological sensorimotor organization offers a different functional picture. Slowly evolving contextual states organize faster trajectories \citep{kiebel2008timescales,hasson2008hierarchy,murray2014intrinsic,chaudhuri2015hierarchical}; goal-directed behavior is structured into tasks, subtasks, and primitive actions \citep{botvinick2009hierarchical}; and continuous experience is segmented at behaviorally meaningful boundaries \citep{zacks2007event,baldassano2017event}. Outcome-sensitive signals evaluate evolving state quality \citep{schultz1997reward}, while internal forward models and probabilistic sensorimotor inference represent expected consequences and uncertainty \citep{kording2004bayesian,wolpert1995sensorimotor,wolpert1998internal,faisal2008noise}. Together, these findings motivate functionally distinct predictive state corresponding to \emph{progress}, \emph{event structure}, and \emph{motor uncertainty}. We use this computational principle as an engineering prior, not as a claim of anatomical equivalence.

Modern VLA research has already begun to expose the first two forms of structure. Progress- and value-aware objectives represent task completion and degraded states \citep{chen2021decision,intelligence2025pi,liu2026steam,zhang2026prts}, while hierarchical and event-conditioned models organize long-horizon behavior through explicit subtasks, stages, or intermediate predictions \citep{belkhale2024rth,zhao2025cot,intelligence2025pi1,intelligence2026pi,xu2026improving,yang2026eventvla}. Uncertainty has followed a different path. Recent VLA systems infer confidence from disagreement between action heads, token entropy, inference-time self-uncertainty, or small policy ensembles \citep{chopra2025everydayvla,choi2026scale,tang2026shifting,romer2026uqvla}. These estimators support monitoring or adaptation, but uncertainty is not learned as a native heteroscedastic output inside the pretrained action expert. The missing capability is therefore a generalist VLA foundation model that learns progress, event intention, and action reliability jointly with action generation, exposes all three online, and uses them inside control.

\begin{figure}[t!]
    \centering
    \includegraphics[width=\linewidth]{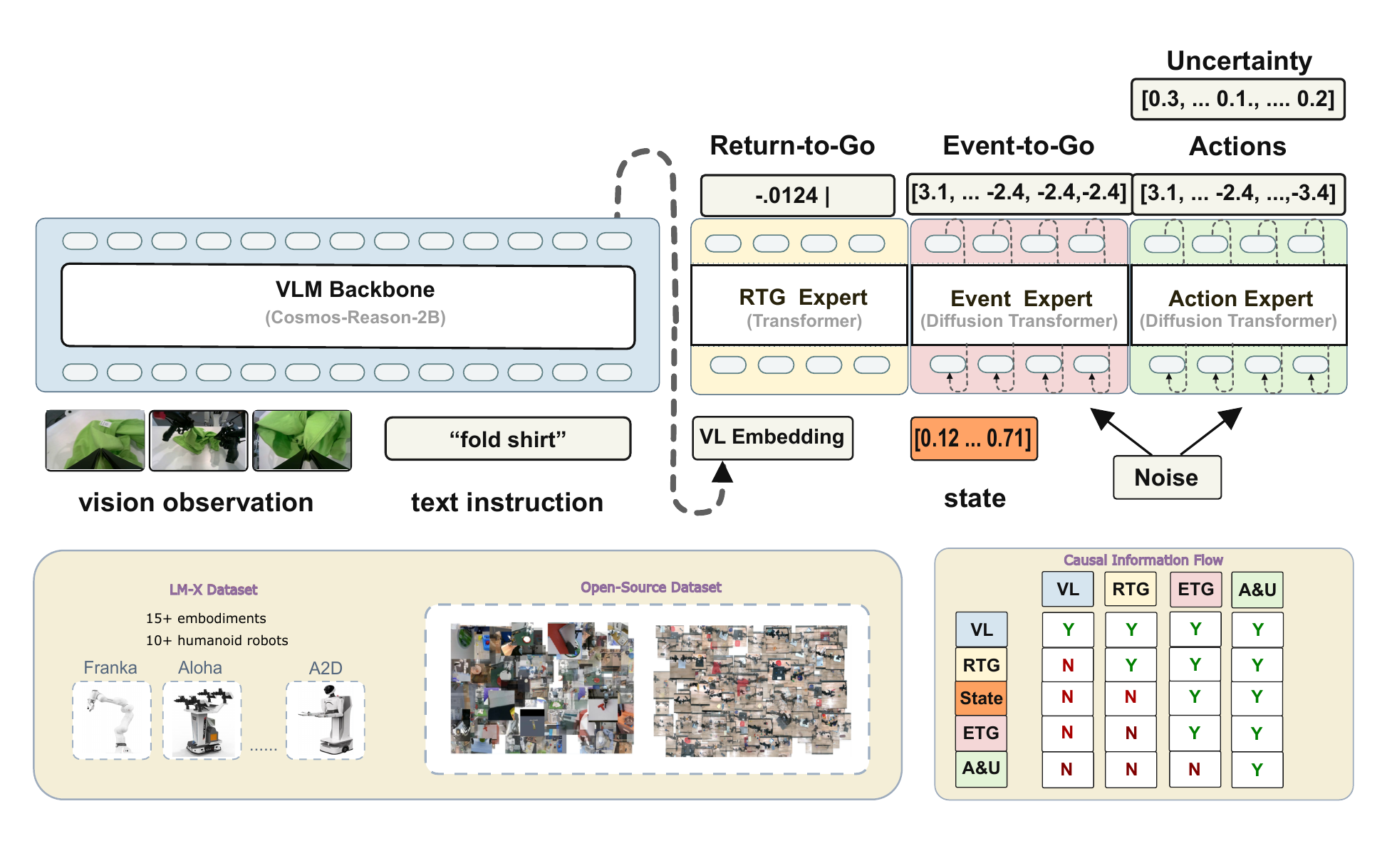}
    \caption{\textbf{Explainable, brain-inspired multi-timescale organization of \method.} Return-to-go (RTG), event-to-go (ETG), and variance explicitly expose progress, intermediate intention, and local action reliability. RTG conditions the event head; its event representation and RTG then condition fine-grained action generation, while uncertainty is estimated inside the action expert. Arrows denote computational conditioning, not anatomical correspondences.}
    \label{fig:architecture}
\end{figure}

We address this gap with \method{} (Figure~\ref{fig:architecture}), an intrinsically explainable generalist VLA foundation model that jointly pretrains action generation with three explicit signals. \emph{Return-to-go (RTG)} is an observation-centric task-scale estimate that answers \emph{is the visible state improving?} \emph{Event-to-go (ETG)} is an intention-centric action chunk that answers \emph{what transition is the policy pursuing?} A heteroscedastic action-flow head predicts motor-scale variance that answers \emph{how reliable is the local command?} RTG conditions ETG, and both condition the fine-grained action expert. To our knowledge, \method{} is the first large-scale generalist VLA to pretrain a native heteroscedastic uncertainty signal inside its action expert. Each signal has a directly supervised operational target, is emitted before action execution, and participates in the action pathway. Explanation is therefore part of the policy's control computation rather than a post-hoc description.

Learning these signals at scale requires more than success-only demonstrations, which show nominal action sequences but rarely expose regression, hesitation, or unreliable states. We augment more than 20,000 hours of heterogeneous real-robot trajectories with outcome, progress, and event labels, including over 1,000 hours of failed policy rollouts. Successful demonstrations define nominal progress and event structure; failures provide missed grasps, stalls, target switches, and other off-nominal states needed to distinguish progress from regression and commitment from ambiguity. Before committing to the costly 20-day pretraining run on 64 NVIDIA B200 GPUs, we use a controlled five-task RoboTwin2.0 study \citep{chen2025robotwin} as a component-attribution gate. Beyond the action-only and single-objective variants, the gate includes an action-pretrained control whose three explanatory heads are instantiated only during downstream adaptation. The complete design raises mean success from 63.6\% for the action-only backbone and 72.8\% for this post-hoc control to 79.6\%, providing matched gate-scale evidence that the three signals are complementary and benefit from participating during pretraining.

After large-scale pretraining, \method{} reaches 74.1\% across 50 randomized-hard RoboTwin2.0 tasks versus 55.4\% for GR00T N1.7 under the same demonstration budget. Across seven real-world tasks on four embodiments, \method{} achieves 73.5\% mean success versus 50.7\% for GR00T N1.7 ($+22.8$ pp), outperforming it on five tasks, tying on one, and trailing on one. Held-out signal evaluation further shows that foundation pretraining raises RTG Spearman correlation from 0.71 to 0.91 and halves its MSE from 0.100 to 0.050; reduces next-event ETG MSE from 0.884 to 0.520; raises action-error AUPRC from 0.49 to 0.59; and extends mean failure-warning lead time from 0.93 to 1.54 seconds.

Our contributions are threefold. (1) We introduce \method{}, a brain-inspired and intrinsically explainable generalist VLA that jointly pretrains RTG, ETG, and action generation and, to our knowledge, learns the first native heteroscedastic uncertainty head inside a large-scale VLA action expert. (2) We develop a scalable outcome, progress, and event annotation pipeline over more than 20,000 hours of robot data, including over 1,000 hours of failed rollouts, together with a cost-aware component-attribution gate that distinguishes joint pretraining from attaching the same heads only downstream before full-scale resources are committed. (3) We show that \method{} outperforms strong generalist baselines in simulation and on real robots, while held-out signal-level evaluation quantitatively tests progress ordering, next-event prediction, action-error detection, and gradient-based failure-warning lead time.

\section{Method}
\label{sec:method}

\subsection{Problem Formulation}
At control step $t$, the robot receives visual observations $o_t$, a language instruction $l$, and proprioceptive state $\mathbf{s}_t$. In addition to a fine-grained action chunk, \method{} emits task progress, an event-level transition, and action uncertainty:
\begin{equation}
    (\hat R_t,\hat{\mathbf{E}}_t,\hat{\mathbf{A}}_t,\hat{\mathbf{q}}_t)
    = \pi_{\Theta}(o_t,l,\mathbf{s}_t),
    \label{eq:policy_interface}
\end{equation}
where $\hat{\mathbf{A}}_t=[\hat{\mathbf{a}}_t,\ldots,\hat{\mathbf{a}}_{t+H-1}]$ has horizon $H=30$, $\hat{\mathbf{E}}_t$ reaches the next annotated semantic boundary, and $\hat{\mathbf{q}}_t$ is the propagated terminal action variance. RTG, ETG, and variance are directly supervised, emitted online, and coupled to the action pathway; the architecture below predicts them in the order progress $\rightarrow$ event $\rightarrow$ action.

\subsection{\method{} Architecture and Information Flow}
\label{sec:architecture}

We build \method{} on the Cosmos-Reason2-2B vision--language backbone \citep{agarwal2025cosmos}, the same backbone family used by GR00T N1.7 \citep{nvidia2026gr00t17}. As shown in Figure~\ref{fig:architecture}, its Qwen-VL front end encodes the instruction and available head, wrist, or embodiment-specific camera views. Proprioception bypasses the vision--language backbone and enters only the event and action experts, separating observation-level task assessment from embodiment-specific control.

Hidden tokens from the 16th backbone layer form the shared representation $\mathbf{z}_t$. A two-layer Transformer implements the RTG expert \citep{vaswani2017attention}; the event and action experts are separate 32-layer Diffusion Transformers \citep{peebles2023scalable} trained with flow matching \citep{lipman2022flow,lipman2024flow,geng2026mean}. The complete model contains approximately 6B parameters.

The RTG expert operates only on $\mathbf{z}_t$. The event expert additionally receives $\mathbf{s}_t$ and RTG, and exposes its hidden state as $\mathbf{z}^{\mathrm{event}}_t$. The action expert then conditions on $\mathbf{z}_t$, $\mathbf{s}_t$, RTG, and $\mathbf{z}^{\mathrm{event}}_t$; its mean-flow and log-variance projections share all preceding features. Ground-truth RTG and event targets supervise this hierarchy during training, while predicted signals are used online. Thus, observation-level progress estimation precedes embodiment-grounded transition prediction, which in turn precedes motor control. Each stage exposes a defined control variable and shapes the next stage without implying an anatomical correspondence.

\begin{figure}[t!]
    \centering
    \includegraphics[width=\linewidth]{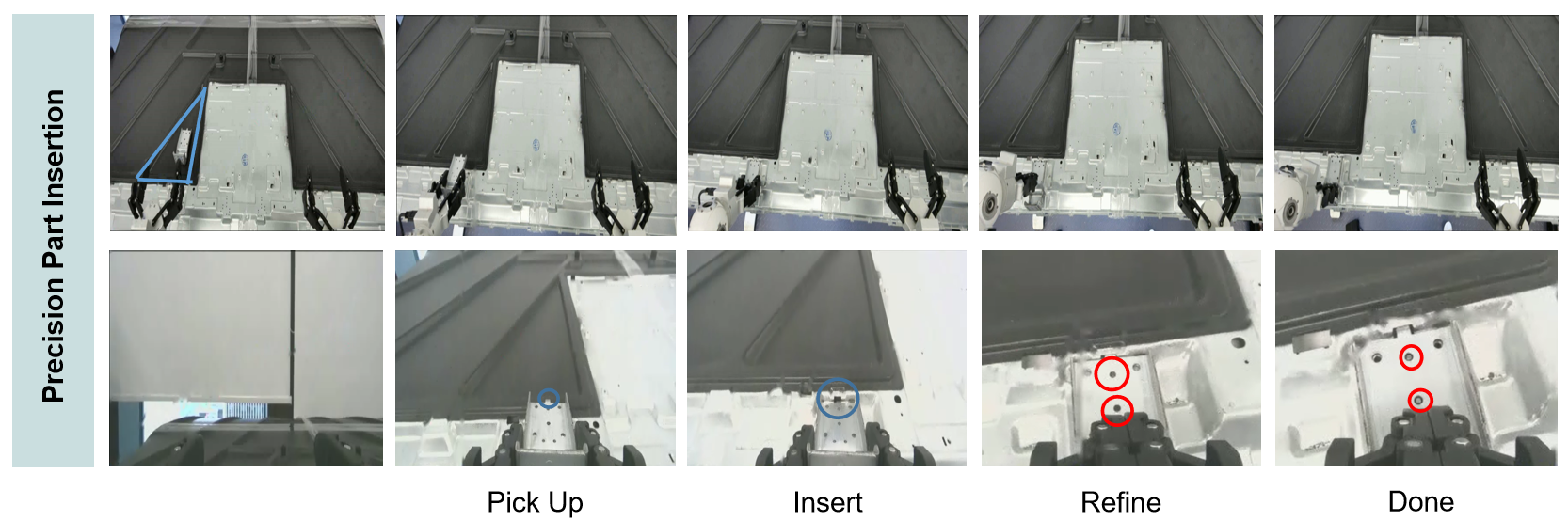}
    \caption{\textbf{Precision part insertion.} From left to right, the robot picks up the part from the blue triangular region, inserts its 3 mm $\times$ 5 mm tip into the slot (blue circles), refines the pose to align two small screw tips with their holes (red circles), and completes the assembly. The sequence combines discrete semantic transitions with millimeter-scale contact.}
    \label{fig:precise_insert}
\end{figure}

ETG is particularly useful when a task contains precise intermediate targets. In such cases, a short local motion is meaningful only in relation to the next contact or alignment milestone. Figure~\ref{fig:precise_insert} illustrates this setting: the same part must pass through distinct pick-up, insertion, and refinement stages before the assembly is complete.

For this task, ETG provides an action-space prior for the next semantic transition, while the action expert resolves the fine local corrections needed to reach it. RTG summarizes whether these transitions advance the task, and variance reports the reliability of the local motion. We next define the three signals in detail.

\subsection{Return-to-Go as Progress and State Quality}
\label{sec:rtg}
Following prior RTG formulations \citep{chen2021decision,zhang2025reinbot,intelligence2025pi}, we define the empirical return-to-go at time $t$ as
\begin{equation}
    R_t = \sum_{t'=t}^{T} r_{t'},
    \label{eq:rtg_definition}
\end{equation}
where $T$ is the terminal timestep. Rewards are constructed from trajectory outcome and remaining duration:
\begin{equation}
    r_t =
    \begin{cases}
        0,  & t=T \text{ and the episode succeeds},\\
        -C, & t=T \text{ and the episode fails},\\
        -1, & \text{otherwise}.
    \end{cases}
    \label{eq:reward}
\end{equation}
Here, $C$ is a large failure constant chosen relative to the maximum successful episode length for the corresponding task. Consequently, RTG approaches zero as a successful trajectory nears completion, while a failed endpoint receives a distinct negative margin. The target combines two signals available at scale: remaining duration along nominal behavior and degraded state quality near observed failure. It is an empirical label on the collected behavior distribution, not the expected return of an optimal policy.

The RTG head models $p_{\phi}(R_t\mid\mathbf{z}_t)$ over $B=128$ ordered bins, providing a bounded representation across tasks of different duration. We recover a scalar estimate from the expected bin center and optimize a smooth $\ell_1$ objective:
\begin{equation}
    \hat R_t = \sum_{b=1}^{B} c_b\,p_{\phi}(b\mid\mathbf{z}_t),
    \qquad
    \mathcal{L}_{\mathrm{RTG}} =
    \operatorname{smooth\text{-}\ell_1}(R_t,\hat R_t),
    \label{eq:rtg_loss}
\end{equation}
where $c_b$ is the center of bin $b$. Ordered bins bound the scalar prediction while the expected-bin regression preserves metric ordering. The RTG head intentionally receives no joint state. This architectural bottleneck encourages it to assess progress from instruction-conditioned scene state rather than memorize embodiment-specific configurations. Proprioception remains available to the event and action heads, which require it for feasible motion generation.

Supervising RTG at every timestep converts a sparse episode outcome into a dense learning signal without requiring hand-designed rewards for individual contacts or object poses. Because an incorrect terminal label or truncated episode shifts every preceding target, we compute RTG only after validating the episode outcome and temporal boundaries.

\subsection{Event-to-Go Prediction}
\label{sec:etg}
Long-horizon manipulation contains sparse, semantically salient states, such as closing the gripper on an object, reaching a pre-insertion pose, establishing tool contact, or completing a fold. These states partition an episode into behaviorally coherent segments. Let $k(t)>t$ be the first verified event index following step $t$. Although an event could be represented as a single target action, action chunking improves behavioral-cloning performance by preserving short-range temporal structure \citep{lazzati2026does,zhao2023act}. Therefore, rather than predicting only one Cartesian pose or end-effector target \citep{agarwal2026cosmos,xu2026improving}, we use an event horizon $H_E$ and construct
\begin{equation}
\mathbf{E}_t=
[\mathbf{a}_t,\ldots,\mathbf{a}_{k(t)},
\underbrace{\mathbf{a}_{k(t)},\ldots,\mathbf{a}_{k(t)}}_{\text{padding to }H_E}],
\label{eq:event_target}
\end{equation}
truncating only when $k(t)-t+1>H_E$. Repeating the terminal event action gives the fixed-length tensor an explicit stopping structure and prevents supervision from leaking into the subsequent subgoal.

The event horizon spans 60 steps, compared with 30 steps for the fine-grained action head, and covers the next event in more than 86\% of training samples. This separation lets ETG represent a semantic transition while the action head focuses on precise local control. Both heads retain compatible action representations, eliminating the need for a separate image generator or hand-designed symbolic skill vocabulary.

The ETG head is trained by conditional flow matching. For flow time $\tau\in[0,1]$, let
\begin{equation}
    \mathbf{E}_t^{\tau}
    = \tau\mathbf{E}_t+(1-\tau)\boldsymbol{\epsilon},
    \qquad
    \boldsymbol{\epsilon}\sim\mathcal{N}(\mathbf{0},\mathbf{I}),
\end{equation}
with target velocity $\mathbf{u}^{\mathrm{event}}=\mathbf{E}_t-\boldsymbol{\epsilon}$. The event loss is
\begin{equation}
    \mathcal{L}_{\mathrm{event}} =
    \mathbb{E}\!\left[
    \left\|
    \mathbf{v}^{\mathrm{event}}_{\psi}
    (\mathbf{E}_t^{\tau},\mathbf{s}_t,\mathbf{z}_t,R_t)
    -\mathbf{u}^{\mathrm{event}}
    \right\|_2^2
    \right].
    \label{eq:event_loss}
\end{equation}
Because ETG is itself an action chunk, it directly represents the motion intended before the next semantic transition. It also supports a useful diagnostic decomposition: an event chunk directed toward the wrong object or pose is consistent with a subgoal-selection error, whereas a suitable event chunk followed by a divergent short action is consistent with a low-level execution error. These interpretations are hypotheses exposed by the interface; causal attribution requires intervention experiments.

\subsection{Uncertainty-Aware Action Flow}
\label{sec:uncertainty_method}
For action chunk $\mathbf{A}_t$, we use the affine probability path
\begin{equation}
\mathbf{A}_t^{\tau}=\tau\mathbf{A}_t+(1-\tau)\boldsymbol{\epsilon},
\quad
\mathbf{u}^{\tau}=\mathbf{A}_t-\boldsymbol{\epsilon},
\quad
\boldsymbol{\epsilon}\sim\mathcal{N}(\mathbf{0},\mathbf{I}).
\label{eq:action_path}
\end{equation}
Standard conditional flow matching regresses a deterministic velocity from $\mathbf{A}_t^{\tau}$ to $\mathbf{u}^{\tau}$ \citep{lipman2023flow}. Multiple base-noise samples can produce different actions, but sample diversity alone does not distinguish valid behavioral multimodality from locally unpredictable control. \method{} therefore predicts an element-wise mean velocity $\bar{\mathbf{v}}_{\theta}^{\tau}$ and log variance $\boldsymbol{\rho}_{\theta}^{\tau}=\log((\boldsymbol{\sigma}_{\theta}^{\tau})^2+\varepsilon)$, conditioned on
\begin{equation}
\mathbf{c}_t^{\tau}=
(\mathbf{A}_t^{\tau},\mathbf{s}_t,\mathbf{z}_t,R_t,
\mathbf{z}^{\mathrm{event}}_t).
\label{eq:action_condition}
\end{equation}

We minimize the Gaussian negative log-likelihood of the target velocity,
\begin{equation}
    \mathcal{L}_{\mathrm{action\text{-}U}}
    = \mathbb{E}\!\left[
        \frac{1}{2}\exp(-\boldsymbol{\rho}_{\theta}^{\tau})
        \left(\bar{\mathbf{v}}_{\theta}^{\tau}(\mathbf{c}_t^{\tau})-
        \mathbf{u}^{\tau}\right)^2
        +\frac{1}{2}\boldsymbol{\rho}_{\theta}^{\tau}
    \right],
    \label{eq:uncertainty_loss}
\end{equation}
where the expectation includes averaging over action dimensions. The residual term rewards accurate velocities, while the log-variance term prevents arbitrary variance inflation. Mean and variance share the action expert and differ only at their final projections, keeping the uncertainty estimate coupled to the features that generate control. This objective estimates conditional residual scale; it does not isolate epistemic uncertainty or certify out-of-distribution safety.

\subsection{Variance Propagation to Action Space}
\label{sec:variance_propagation}
The flow head predicts local velocity variance, whereas online monitoring requires uncertainty in the final action sample. We therefore propagate a diagonal approximation through numerical integration. For a fixed base-noise sample, we initialize $\mathbf{q}^{0}=\mathbf{0}$ so that the trace measures accumulated predictive variance rather than dispersion across base samples. Under a forward Euler update,
\begin{equation}
    \bar{\mathbf{x}}^{\tau+\delta}
    =\bar{\mathbf{x}}^{\tau}
    +\bar{\mathbf{v}}^{\tau}\delta,
\end{equation}
the element-wise variance evolves approximately as
\begin{equation}
    \operatorname{Var}(\mathbf{x}^{\tau+\delta})
    \approx \operatorname{Var}(\mathbf{x}^{\tau})
    +(\boldsymbol{\sigma}^{\tau}\delta)^2
    +2\delta\operatorname{Cov}(\mathbf{x}^{\tau},\mathbf{v}^{\tau}).
    \label{eq:variance_update}
\end{equation}
The covariance term captures how uncertainty already present in the flow state changes the predicted mean velocity. We approximate its diagonal with a first-order Taylor expansion and Hutchinson's estimator \citep{han2026flow}:
\begin{equation}
    \operatorname{Cov}(\mathbf{x}^{\tau},\mathbf{v}^{\tau})
    \approx \frac{1}{N}\sum_{i=1}^{N}
    (\sqrt{\mathbf{q}^{\tau}}\circ\boldsymbol{\epsilon}_i)
    \circ
    \mathbf{J}^{\tau}(\bar{\mathbf{x}}^{\tau})
    (\sqrt{\mathbf{q}^{\tau}}\circ\boldsymbol{\epsilon}_i),
    \label{eq:covariance}
\end{equation}
where $\mathbf{q}^{\tau}=\operatorname{Var}(\mathbf{x}^{\tau})$, $\boldsymbol{\epsilon}_i$ is a Rademacher vector, $\circ$ denotes element-wise multiplication, and $\mathbf{J}^{\tau}$ is the Jacobian of the mean velocity with respect to the flow state. This is a first-order diagonal approximation; cross-dimensional covariance is not retained.

The scalar uncertainty used in Section~\ref{sec:uncertainty_analysis} is the mean terminal variance across the selected action dimensions:
\begin{equation}
    U_t=\frac{1}{d}\sum_{j=1}^{d}
    \operatorname{Var}(x_{t,j}^{\tau=1}).
    \label{eq:uncertainty_score}
\end{equation}
Larger $U_t$ denotes greater modeled action dispersion. Arm-specific traces average only dimensions assigned to the corresponding arm, allowing uncertainty to be localized in bimanual execution. All reported rollouts use the mean velocity without uncertainty guidance, so the traces are diagnostic outputs rather than the consequence of an uncertainty-minimizing controller.

\subsection{Joint Training and Online Interpretation}
The complete pretraining objective is
\begin{equation}
    \mathcal{L}_{\mathrm{LM\text{-}X}}
    =\lambda_R\mathcal{L}_{\mathrm{RTG}}
    +\lambda_E\mathcal{L}_{\mathrm{event}}
    +\lambda_A\mathcal{L}_{\mathrm{action\text{-}U}},
    \label{eq:global_loss}
\end{equation}
with fixed loss weights $\lambda_R,\lambda_E,$ and $\lambda_A$. Large-scale pretraining jointly optimizes the shared backbone, all three predictive-state heads, and action generation on a weighted mixture of robot data. The explanatory interface is therefore learned with the foundation policy rather than appended to a frozen black-box VLA during downstream adaptation. Task-specific post-training subsequently adapts the complete model to precise-contact, long-horizon, and deformable-object tasks using curated demonstrations, while the predictive-state heads remain active throughout adaptation and inference.

At inference time, \method{} estimates progress from the current visual--language representation, predicts the next event conditioned on progress and proprioception, and integrates a fine-grained action flow conditioned on both signals. All outputs are available at every control step. A downstream monitor can therefore test for decreasing RTG, event changes, or variance spikes before terminal failure. We evaluate whether these signals correspond to recognizable execution dynamics; using them to trigger recovery is left to future work.

\section{Dataset and Annotation Pipeline}
\label{sec:data}

\subsection{Data Representation and Composition}
Each labeled timestep is represented as
\begin{equation}
    (o_t,l,\mathbf{s}_t,R_t,\mathbf{E}_t,\mathbf{A}_t),
    \label{eq:data_tuple}
\end{equation}
This representation augments the conventional observation--instruction--state--action tuple with progress and next-event targets. The training mixture contains more than 20,000 hours of real-robot trajectories, including over 1,000 hours of failed policy rollouts generated by ACT \citep{zhao2023act}, Diffusion Policy \citep{chi2023diffusion}, $\pi$-series, and GR00T-based agents. We convert both public datasets and newly collected trajectories to this common schema.

The full language instruction is retained at every timestep. Prompts for multi-object and bimanual tasks specify object attributes, spatial relations, orientations, and arm assignments. For example, an instruction may require the left arm to place a white paper ball into a bin while the right arm places a red pen into a holder. Such detail is necessary because the same scene may admit multiple locally feasible actions, whereas progress and event predictions are defined relative to the complete task.

The data span single-arm, dual-arm, and humanoid embodiments with heterogeneous morphologies and sensing configurations. To encourage cross-embodiment transfer, the progress head does not receive joint state; the event and action heads retain proprioception to generate kinematically feasible motions.

\subsection{Outcome and RTG Labels}
Each episode receives a terminal success or failure label. Successful episodes accumulate the ordinary step penalty until completion, whereas failed episodes additionally receive the terminal penalty in Eq.~\eqref{eq:reward}. We compute RTG backward from the terminal state, converting a single outcome label into a dense target that preserves the temporal ordering of nominal progress.

Failed rollouts broaden the state distribution beyond teleoperated expert behavior, exposing the model to missed grasps, target switches, hesitation, occlusion, and other off-nominal states. They also discourage the trivial association between elapsed time and task progress.

\subsection{Event Definition and Verification}
We define an event as a semantically meaningful intermediate state that partitions a trajectory into interpretable subgoals. Annotation proceeds in two stages: task-dependent signals first propose candidate timesteps, after which human annotators verify each candidate and correct boundaries that do not match the intended transition. This procedure avoids frame-by-frame labeling while preserving task-specific semantics.

Event proposals reflect the physical structure of each task family. Gripper transitions are informative for object transfer; explicit alignment markers identify insertion boundaries; velocity minima reveal stabilization during deformable-object manipulation; and contact signals combined with motion reversals delimit wiping strokes. Despite these different proposal mechanisms, every annotation yields the same target: the action sequence from the current timestep to the next verified event. Appendix~\ref{app:event_define} provides details.

\section{Experiments}
\label{sec:experiments}

\subsection{Evaluation Questions and Metrics}
Our experiments address five questions aligned with the predictive-state and explainability hypotheses:
\begin{enumerate}
    \item Before costly full-scale pretraining, do the three explicit signals contribute complementary gains, and does joint gate-scale pretraining outperform attaching the same heads only downstream?
    \item Does the complete factorization transfer across a broad simulation task distribution?
    \item Do the gains persist across real robots and manipulation regimes?
    \item Do explicit RTG and ETG quantitatively recover task progress and the next verified event?
    \item Is explicit action variance aligned with local prediction error, and can its temporal gradient warn of impending failure?
\end{enumerate}

Full pretraining takes approximately 20 days on 64 NVIDIA B200 GPUs, making late architectural revisions expensive and a second matched foundation-scale run impractical. We therefore separate component attribution from system-scale evaluation. At gate scale, every variant uses the same 45 RoboTwin2.0 pretraining tasks, five disjoint downstream tasks, optimization budget, and evaluation protocol. In addition to the action-only and single-head models, an \emph{action-pretrained + post-hoc heads} control first learns the action-only objective on the 45 pretraining tasks and instantiates RTG, ETG, and uncertainty only during downstream post-training. The jointly pretrained \method{} uses the same final architecture and downstream supervision, but exposes the representation to all three objectives during pretraining. This controlled comparison attributes gate-scale differences to when the explanatory objectives enter training; the later 20,000-hour experiment evaluates the selected system rather than repeating component attribution at foundation scale.

After the gate supports the complete design, we pretrain \method{} with all predictive-state heads on more than 20,000 hours of real-robot data; Appendix~\ref{app:train_recipe} provides additional details.

Our primary control metric is binary task success, with absolute differences reported in percentage points. The pretraining-gate study weights its five held-out tasks equally. For the full RoboTwin2.0 benchmark, each model is post-trained with 50 randomized-hard demonstrations per task and evaluated over 100 trials per task; the aggregate score weights all 50 tasks equally. For each real-world task, we collect 500--1,200 post-training episodes and evaluate all methods under the same physical setup and success criterion. A separate held-out set of 50 episodes per task supports signal-level evaluation. RTG quality is measured by trajectory-wise Spearman correlation and MSE against normalized progress targets, while ETG quality is measured by masked next-event MSE in normalized action coordinates. Action uncertainty is evaluated by AUPRC for detecting high-error chunks. Finally, we record 10 failure episodes per task and compute warning lead time from the temporal uncertainty gradient, using a threshold that yields a 10\% false-alarm rate on successful validation trajectories.

For every real-world signal-level study, we compare two architecture-matched variants using identical task-specific data and optimization. We denote them as \emph{w/o pretraining}, which initializes the complete backbone and three-head architecture from scratch and learns only from task-specific post-training data, and \emph{w/ pretraining}, which initializes the same architecture from the 20,000-hour jointly pretrained checkpoint before applying the identical post-training recipe. This is a transfer evaluation rather than a component-attribution ablation: it measures the benefit of foundation pretraining under a matched downstream setup. 

Open-source release can be found at \url{https://github.com/loongOpen/LoongWu-LM-X-VLA}

\subsection{Controlled Component Attribution before Full Pretraining}
\label{sec:ablation}
This experiment is deliberately conducted \emph{before} large-scale pretraining: its purpose is to provide the component-level evidence needed to finalize the architecture before initiating the 20-day, 64-B200 run. We isolate each objective in a controlled RoboTwin2.0 study \citep{chen2025robotwin}. Five of the benchmark's 50 tasks are excluded from gate pretraining and reserved for downstream post-training and evaluation; Appendix~\ref{app:ablation} details the setup. The \emph{Backbone} uses deterministic flow matching without RTG or event supervision. \emph{LM-RTG}, \emph{LM-Event}, and \emph{LM-U} add one component at a time. The \emph{post-hoc heads} control starts from the action-only pretrained backbone and learns all three heads only on the five downstream tasks, whereas \method{} includes all three objectives throughout gate pretraining. The latter pair has the same downstream architecture, annotations, data, and optimization, making it the gate's primary attribution comparison.

\begin{table}[H]
    \caption{Controlled component-attribution gate on five held-out RoboTwin2.0 tasks (success rate, \%). Post-hoc starts from the action-only pretrained backbone and learns all three explanatory heads only downstream; LM-X jointly pretrains them. The final row macro-averages tasks.}
    \label{tab:ablation}
    \centering
    \small
    \resizebox{\linewidth}{!}{%
    \begin{tabular}{lrrrrrr}
        \toprule
        Task & Backbone & LM-RTG & LM-Event & LM-U & Post-hoc & LM-X \\
        \midrule
        Handover microphone & 76 & 86 & \textbf{92} & 82 & 82 & 88 \\
        Lift pot & 84 & 90 & 86 & \textbf{94} & 86 & 92 \\
        Open microwave & 54 & \textbf{68} & 16 & 54 & 62 & 62 \\
        Rank RGB blocks & 48 & 38 & 66 & 44 & 74 & \textbf{90} \\
        Hit block with hammer & 56 & 62 & 60 & 52 & 60 & \textbf{66} \\
        \midrule
        Average & 63.6 & 68.8 & 64.0 & 65.2 & 72.8 & \textbf{79.6} \\
        \bottomrule
    \end{tabular}%
    }
\end{table}

\begin{figure}[t]
    \centering
    \small
    \setlength{\tabcolsep}{3pt}
    \begin{tabular}{@{}lcr@{}}
        Action-only backbone & {\color{gray!70}\rule{3.18cm}{6pt}} & 63.6 \\
        Strongest single head & {\color{gray!70}\rule{3.44cm}{6pt}} & 68.8 \\
        Action-pretrained + post-hoc heads & {\color{gray!70}\rule{3.64cm}{6pt}} & 72.8 \\
        Jointly pretrained \method{} & {\color{blue!65}\rule{3.98cm}{6pt}} & \textbf{79.6}
    \end{tabular}
    \caption{Gate-scale attribution summary (macro success, \%). Joint pretraining exceeds both the strongest isolated objective and the matched post-hoc-head control.}
    \label{fig:gate_attribution_summary}
\end{figure}

The gate reveals that no single component dominates across tasks (Table~\ref{tab:ablation}). RTG performs best on opening the microwave, improving over the backbone by 14 points. Event prediction yields the largest gain on microphone handover (+16), where the transfer provides a clear intermediate boundary. Uncertainty performs best on bimanual pot lifting (+10), which admits substantial variation in motion and object stability. Thus, each objective can provide useful structure, but its value is task dependent.

The gate also reveals that isolated auxiliary objectives can be brittle before full-scale pretraining. In the most pronounced case, event-only training reduces open-microwave success from 54\% to 16\%, indicating that coarse event guidance alone may be insufficient for contact-intensive action sequences. Nevertheless, event supervision yields substantial gains on block ranking and microphone handover. To preserve these benefits while mitigating over-reliance on event priors, we drop $\mathbf{z}^{\mathrm{event}}_t$ from the action expert's input with a probability of 20\% during pretraining, encouraging the action DiT to remain effective without event conditioning. During post-training, we treat the event embedding as a task-specific hyperparameter selected on downstream validation data. It conditions the action expert for block ranking and hammering in both the Post-hoc and LM-X variants in Table~\ref{tab:ablation}, and is omitted for the other three gate tasks. Appendix~\ref{app:add_event} reports the corresponding ablation.

For the pretraining decision, the complete model is markedly more consistent: LM-X improves all five tasks over the backbone, with gains of 12 points for handover, 8 for both pot lifting and microwave opening, 42 for block ranking, and 10 for hammering. Overall, LM-X raises mean success from 63.6\% to 79.6\% (+16.0 points), whereas RTG, event prediction, and uncertainty alone improve the mean by 5.2, 0.4, and 1.6 points, respectively. The complete model also exceeds the strongest single-component variant by 10.8 points. More importantly for attribution, attaching the three heads only after action pretraining reaches 72.8\%; exposing the same final architecture to the three objectives during gate pretraining provides a further 6.8-point gain (Figure~\ref{fig:gate_attribution_summary}).

The single-component comparisons show complementarity, while the matched post-hoc control asks a separate question: whether the same supervision can be deferred until downstream adaptation. ETG supplies a coarse target, RTG evaluates the resulting state relative to completion, and uncertainty captures ambiguity in converting the target into fine-grained control. On block ranking, for example, RTG and uncertainty individually underperform the backbone, yet their joint use with ETG reaches 90\%, exceeding the backbone by 42 points and the post-hoc control by 16 points. These gate-scale results do not identify the underlying representation geometry, but they directly support jointly introducing the three objectives before resources are committed to the single full-scale pretraining run.

\subsection{RoboTwin2.0 Benchmark}
\label{sec:simulation}
We next evaluate generalization across all 50 RoboTwin2.0 tasks on the Aloha-AgileX embodiment. Each model is post-trained with 50 demonstrations per task from the randomized-hard split and evaluated over 100 trials per task. Because LM-X pretraining contains neither RoboTwin2.0 nor other simulation data, this benchmark measures adaptation of a real-robot-pretrained representation to a new visual and dynamical domain.

Throughout the experiments, GR00T denotes the public GR00T N1.7 checkpoint and code snapshot available in July 2026 \citep{nvidia2026gr00t17}, not the GR00T N1.0 model described in the original paper \citep{bjorck2025gr00tn1}. We initialize N1.7 from its released weights and post-train it on the same 50 downstream tasks. Because the pretraining mixtures and parameterizations differ, this is an end-model comparison rather than a controlled architecture or pretraining-data ablation. Both systems receive the same downstream demonstration budget and use the same randomized-hard evaluation protocol.

\begin{table}[t]
    \caption{RoboTwin2.0 success rate (\%). We show representative tasks and the mean across all 50 randomized-hard tasks.}
    \label{tab:simulation}
    \centering
    \small
    \begin{tabular}{lrr}
        \toprule
        Task & GR00T N1.7 & \method{} \\
        \midrule
        \texttt{adjust\_bottle} & 98 & \textbf{100} \\
        \texttt{beat\_block\_hammer} & \textbf{40} & 25 \\
        \texttt{blocks\_ranking\_rgb} & 60 & \textbf{93} \\
        \texttt{blocks\_ranking\_size} & 52 & \textbf{82} \\
        \texttt{click\_alarmclock} & \textbf{100} & \textbf{100} \\
        \multicolumn{3}{c}{$\cdots$} \\
        \texttt{stack\_blocks\_two} & 72 & \textbf{97} \\
        \texttt{stack\_bowls\_three} & 63 & \textbf{75} \\
        \texttt{stack\_bowls\_two} & 91 & \textbf{97} \\
        \texttt{stamp\_seal} & 22 & \textbf{74} \\
        \texttt{turn\_switch} & 61 & \textbf{79} \\
        \midrule
        Average over all 50 tasks & 55.4 & \textbf{74.1} \\
        \bottomrule
    \end{tabular}
\end{table}

LM-X improves the 50-task mean from 55.4\% to 74.1\%, an absolute gain of 18.7 points. Among the ten representative tasks in Table~\ref{tab:simulation}, LM-X outperforms GR00T N1.7 on eight, ties on one, and underperforms on one. This post-pretraining result covers the complete benchmark rather than only the five tasks used by the earlier pretraining gate. Appendix~\ref{app:sim_result} reports all per-task results.

The largest representative gain occurs on \texttt{stamp\_seal}, where success rises from 22\% to 74\% (+52 points). This task requires approach, contact alignment, and forceful execution, with errors at any stage invalidating the trial. LM-X also improves color- and size-based block ranking by 33 and 30 points, respectively, and two-block stacking by 25 points. These tasks require instruction-dependent sequencing across multiple interactions. While the aggregate comparison cannot attribute gains to individual heads, the pattern is consistent with the pretraining-gate ablation that motivated retaining the structured objectives: intermediate supervision is particularly useful for multi-stage tasks.

Several simple tasks are already saturated: both methods reach 100\% on clicking the alarm clock, and bottle adjustment improves only from 98\% to 100\%. The aggregate gain is therefore concentrated on more challenging tasks rather than distributed uniformly across the benchmark.

The improvement is not universal. LM-X trails GR00T by 15 points on \texttt{beat\_block\_hammer}, despite improving over the backbone on the related held-out hammer task in the pretraining-gate ablation (Table~\ref{tab:ablation}). Differences in downstream data, task configuration, or contact dynamics may explain this reversal across evaluation settings. This result motivates repeated training seeds and per-task confidence intervals in future evaluations.

\subsection{Real-World Evaluation}
\label{sec:real_world}

\begin{table}[h]
\centering
\caption{Real-world success rate (\%). $\Delta_{\mathrm{G}}$ is the \method{} improvement over GR00T N1.7 in percentage points (pp).}
\label{tab:real_world_results}
\small
\setlength{\tabcolsep}{4.5pt}
\begin{tabular}{@{}p{0.46\linewidth}rrrr@{}}
\toprule
Task & GR00T N1.7 & $\pi_{0.5}$ & \method{} & $\Delta_{\mathrm{G}}$ \\
\midrule
Tape-roll pick-and-place & \textbf{75} & 45 & 65 & $-10$ \\
Precision part insertion & 20 & 0 & \textbf{80} & $+40$ \\
Tableware organization & 20 & 15 & \textbf{90} & $+70$ \\
Cloth folding & 60 & 30 & \textbf{80} & $+20$ \\
Battery insertion & 35 & \textbf{50} & 40 & $+5$ \\
Water-bottle placement, dual-arm & 75 & 40 & \textbf{90} & $+15$ \\
Water-bottle placement, single-arm & 70 & \textbf{75} & 70 & $0$ \\
\midrule
Mean & 50.7 & 36.4 & \textbf{73.5} & $\mathbf{+22.8}$ \\
\bottomrule
\end{tabular}
\end{table}

We evaluate seven tasks on four real embodiments. Tape-roll pick-and-place runs on AgileX-Aloha; precision insertion, tableware organization, and cloth folding run on Astribot S1; battery insertion runs on Tianji-Marvin; and the two water-bottle tasks run on Loong S1. During task-specific post-training, $\mathbf{z}^{\mathrm{event}}_t$ conditions the LM-X action expert for precision insertion, tableware organization, and dual-arm water-bottle placement; validation selects the configuration without event conditioning for the other four tasks. Each task uses 20 trials, with identical task-specific data and success criteria across methods (Appendix~\ref{app:real_setting}). In addition to GR00T N1.7, we include $\pi_{0.5}$ \citep{intelligence2025pi1}, a strong generalist VLA that combines heterogeneous co-training with high-level semantic prediction. We adapt all three policies with the same downstream demonstration budget, observations, instructions, success criteria, and evaluation trials. Because their pretraining mixtures and parameterizations differ, these are end-model comparisons rather than controlled pretraining ablations.

These platforms are deliberately low-frequency or unseen in the pretraining mixture. AgileX-Aloha and Astribot S1 each account for 4.22\% of the data, Tianji-Marvin accounts for 0.66\%, and Loong S1 is absent. The tests are not zero-shot because every model receives task-specific post-training, but the limited embodiment overlap makes it less likely that the results are driven by repeated exposure to a dominant robot morphology. In particular, the two Loong S1 tasks test whether the pretrained representation can transfer to a previously unseen embodiment.

\method{} achieves 73.5\% mean success versus 50.7\% for GR00T N1.7 ($+22.8$ pp) and 36.4\% for $\pi_{0.5}$ ($+34.3$ pp). Relative to GR00T, it performs better on five tasks, ties on one, and trails on one; the median task-wise gain is $+15$ pp. The largest improvement occurs on tableware organization ($+70$ pp), which contributes 14 of the 28 net additional successes. Excluding this task, \method{} still leads by $+11.7$ pp (67.5\% vs.\ 55.8\%), so the aggregate advantage does not depend on a single task. Across the three Astribot S1 behaviors, mean success rises from 33.3\% to 85.0\%; across the two Loong S1 tasks, it rises from 72.5\% to 80.0\%. Tape-roll placement is the only regression relative to GR00T, and battery insertion remains the most difficult task for \method{} at 40\%. Precision part insertion (Figure~\ref{fig:precise_insert}) reaches 80\% despite requiring a constrained insertion followed by millimeter-scale alignment of two small screw tips. 

\subsection{RTG Tracks Progress and Local Regressions}
\label{sec:rtg_analysis}
Ground-truth RTG and predictions are linearly normalized using task-specific training-set statistics. We compute Spearman correlation within each trajectory before macro-averaging across trajectories and tasks; MSE additionally tests numerical fidelity. Table~\ref{tab:rtg_pretrain_compare} evaluates signal transfer from foundation pretraining against learning the complete architecture only from post-training data.

\begin{table}[H]
\centering
\caption{RTG quality across seven real-world tasks. Both variants use the same complete architecture and post-training recipe; they differ only in whether training starts from the jointly pretrained checkpoint.}
\label{tab:rtg_pretrain_compare}
\small
\setlength{\tabcolsep}{4.2pt}
\begin{tabular}{@{}lrrrr@{}}
\toprule
& \multicolumn{2}{c}{Progress Spearman $\rho\uparrow$} & \multicolumn{2}{c}{MSE on test set$\downarrow$} \\
\cmidrule(lr){2-3}\cmidrule(lr){4-5}
Task & \shortstack{w/o\\pretraining} & \shortstack{w/\\pretraining} & \shortstack{w/o\\pretraining} & \shortstack{w/\\pretraining} \\
\midrule
Pick and place tape roll & 0.73& 0.86& 0.087 & 0.033 \\
Precision part insertion & 0.78& 0.97& 0.085 & 0.048 \\
Tableware organization & 0.75& 0.94& 0.118 & 0.068 \\
Cloth folding & 0.70& 0.91& 0.070 & 0.044\\
Battery insertion & 0.56& 0.79& 0.019 & 0.011 \\
Bottle placement, dual-arm & 0.71& 0.93& 0.127 & 0.051 \\
Bottle placement, single-arm & 0.72& 0.95& 0.193 & 0.094 \\
\midrule
Average & 0.71 & \textbf{0.91} & 0.100 & \textbf{0.050} \\
\bottomrule
\end{tabular}
\end{table}

\begin{figure*}[t]
    \centering
    \begin{subfigure}[t]{0.24\textwidth}
        \includegraphics[width=\linewidth]{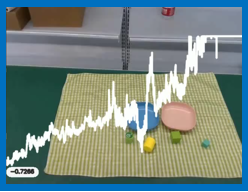}
        \caption{Start}
    \end{subfigure}
    \begin{subfigure}[t]{0.24\textwidth}
        \includegraphics[width=\linewidth]{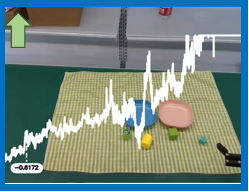}
        \caption{Right gripper appears}
    \end{subfigure}
    \begin{subfigure}[t]{0.24\textwidth}
        \includegraphics[width=\linewidth]{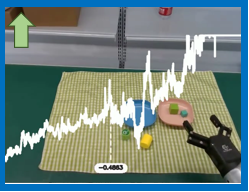}
        \caption{Subtask halfway done}
    \end{subfigure}
    \begin{subfigure}[t]{0.24\textwidth}
        \includegraphics[width=\linewidth]{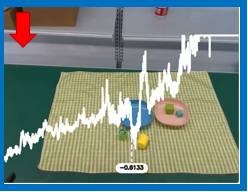}
        \caption{Gripper out of view}
    \end{subfigure}

    \vspace{0.5em}

    \begin{subfigure}[t]{0.24\textwidth}
        \includegraphics[width=\linewidth]{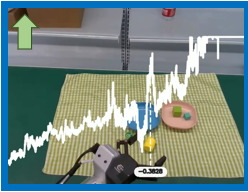}
        \caption{Correctly aligned}
    \end{subfigure}
    \begin{subfigure}[t]{0.24\textwidth}
        \includegraphics[width=\linewidth]{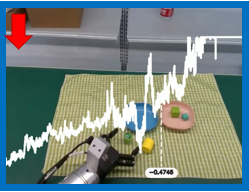}
        \caption{Missed grasp, switches target}
    \end{subfigure}
    \begin{subfigure}[t]{0.24\textwidth}
        \includegraphics[width=\linewidth]{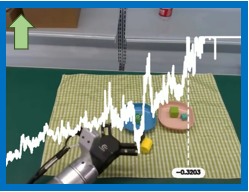}
        \caption{Third object placed}
    \end{subfigure}
    \begin{subfigure}[t]{0.24\textwidth}
        \includegraphics[width=\linewidth]{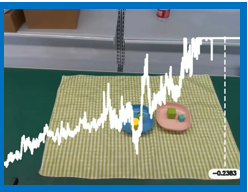}
        \caption{Done}
    \end{subfigure}

    \caption{RTG visualization during a held-out bimanual sorting episode. The instruction is to place the green and blue objects to the right of the pink plate using the right hand, and to place the green and yellow objects to the left of the blue plate using the left hand.}
    \label{fig:rtg_astroid}
\end{figure*}

With foundation pretraining, average progress correlation increases from $0.71$ to $0.91$ and average MSE falls from $0.100$ to $0.050$. Because both variants share the architecture, post-training data, and optimization, the comparison directly measures signal transfer from the foundation checkpoint. It does not isolate which pretraining loss causes the gain; the controlled gate in Section~\ref{sec:ablation} addresses that question. Figures~\ref{fig:rtg_astroid} and~\ref{fig:rtg_pears} next show how the pretrained signal evolves within held-out episodes. Under Eq.~\eqref{eq:reward}, a higher (less negative) RTG indicates that the observed state is estimated to be closer to successful completion. A meaningful progress signal should therefore rise over a successful trajectory while remaining responsive to local regressions; elapsed time alone could explain only the first behavior.

Figure~\ref{fig:rtg_astroid} shows a multi-object, bimanual task. RTG first rises when the right gripper enters the head-camera view in panel (b), indicating that the progress head associates visible task engagement with improved state quality. A larger increase follows completion of the first half of the instruction in panel (c), linking the score to semantic task progress rather than motion alone.

Local decreases provide stronger evidence. RTG falls when the robot returns toward its initial pose and both grippers leave the view in panel (d), then recovers as the left gripper approaches the yellow object in panel (e). In panel (f), the controller abandons the yellow object without grasping it and redirects toward the green object; RTG decreases immediately rather than only at episode termination. This non-monotonic response is inconsistent with a simple elapsed-time heuristic and reflects a visible execution regression.

The score resumes its upward trend as the remaining placements are completed. The trace thus combines a global increase toward completion, stepwise changes near semantic events, and rapid decreases at visible anomalies---properties that can help identify when a long-horizon rollout stops improving.
\begin{figure}[t]
    \centering
    \includegraphics[width=\linewidth]{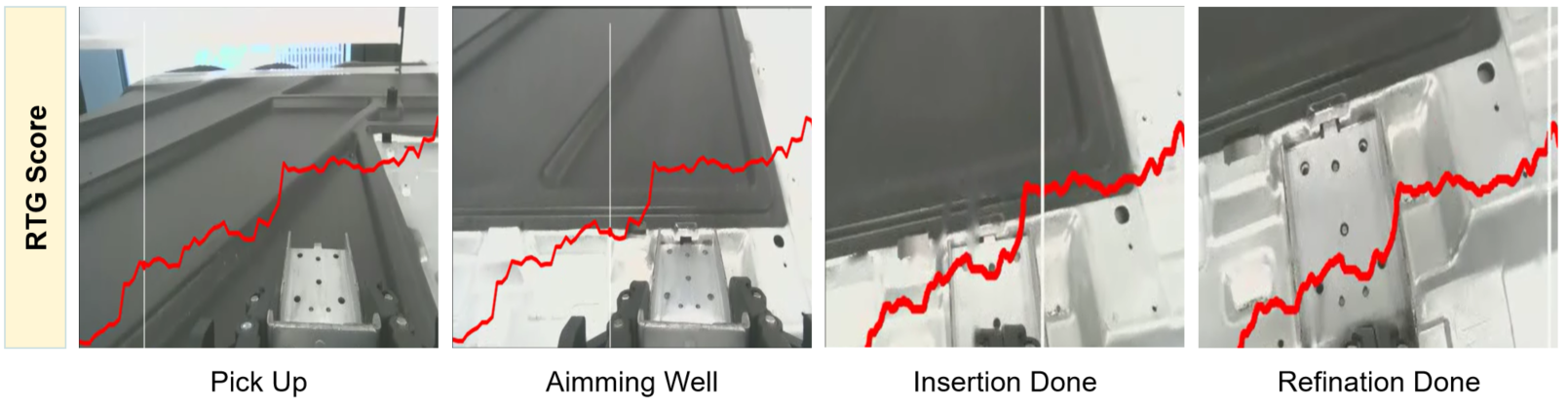}
    \caption{RTG during precision part insertion. From left to right, a successful grasp produces the first large increase; the middle two panels bracket insertion and show a clear boost after the tip enters the slot; the final increase follows pose refinement, when the two small screw tips enter their holes. White cursors mark the illustrated timesteps.}
    \label{fig:rtg_hard}
\end{figure}

Figure~\ref{fig:rtg_hard} provides a contact-rich counterpart to the sorting example. The four snapshots bracket three verified state changes. In the first panel, the robot has secured the part, and RTG shows its first large increase. The second and third panels are immediately before and after the 3 mm $\times$ 5 mm tip enters the slot; the pronounced boost between them indicates that the score recognizes successful insertion rather than generic approach motion. The last rise occurs after refinement, when the two small screw tips are seated in their corresponding holes and the assembly is complete. These unequal, event-aligned steps are difficult to explain as an elapsed-time signal. Appendix~\ref{app:rtg_fail_case} presents a complementary case in which perceptual ambiguity lowers RTG despite continued task completion.

\subsection{ETG Predicts the Next Semantic Event}
\label{sec:etg_analysis}

ETG is not evaluated as a categorical subtask label. Its prediction is an embodiment-native action sequence that should terminate at the next verified semantic boundary. For every held-out query, we compute MSE in normalized action coordinates over the valid, unpadded portion of the target sequence from the current state to the next event. This next-event MSE evaluates both the predicted path and its event-scale horizon while respecting variable event duration. Table~\ref{tab:etg_pretrain_compare} evaluates signal transfer from foundation pretraining against learning the complete architecture only from post-training data.

\begin{figure}[htp]
    \centering
    \includegraphics[width=\linewidth]{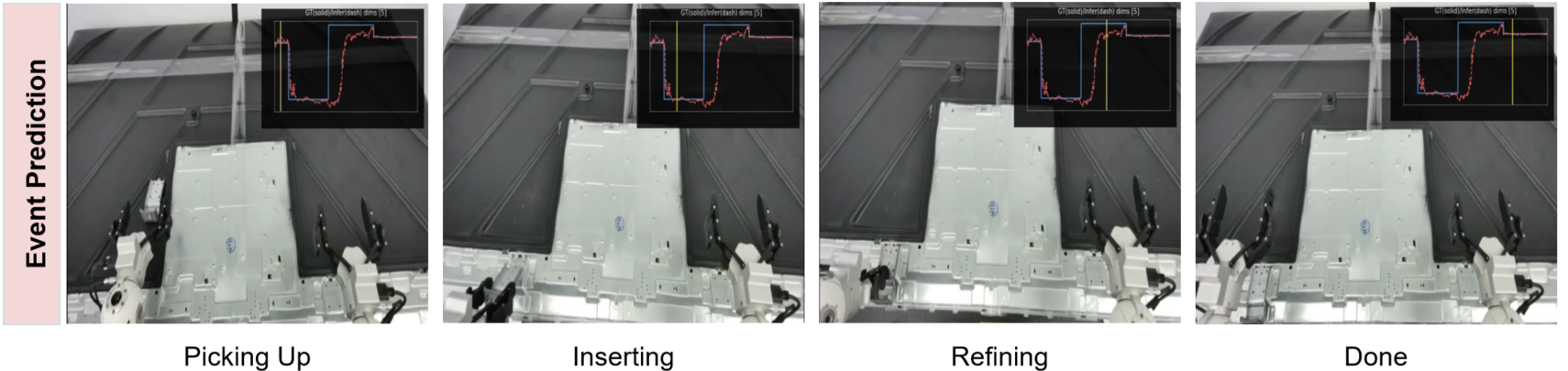}
    \caption{ETG prediction during precision part insertion. The upper-right inset of each panel overlays the event ground truth (GT; blue) and prediction (red) on validation trajectories. We show the wrist-joint dimension with the largest target variation. Each step-like transition of the blue target denotes a new event, and the prediction closely follows both its value and timing, providing an action-space prior for the next event across pick-up, insertion, refinement, and completion.}
    \label{fig:etg_insert}
\end{figure}

Figure~\ref{fig:etg_insert} visualizes the event-scale structure measured by next-event MSE. For each validation example, the inset selects the wrist-joint action dimension with the largest ground-truth change and compares the target trajectory with the ETG prediction. The red prediction follows the blue target as it switches between pick-up, insertion, refinement, and completion. ETG therefore does more than recognize the current stage: its action sequence gives the downstream expert a timely, embodiment-native prior for the motion leading to the next verified event.

\begin{table}[H]
\centering
\caption{Next-event ETG MSE across seven real-world tasks. Both variants use the same architecture, event annotations, and post-training recipe; they differ only in whether training starts from the jointly pretrained checkpoint. Lower is better.}
\label{tab:etg_pretrain_compare}
\small
\setlength{\tabcolsep}{7pt}
\begin{tabular}{@{}lrr@{}}
\toprule
Task & w/o pretraining & w/ pretraining \\
\midrule
Pick and place tape roll & 2.313 & 1.634 \\
Precision part insertion &  0.298 & 0.170 \\
Tableware organization & 0.355 & 0.183 \\
Cloth folding & 1.226& 0.503\\
Battery insertion & 0.439 & 0.235 \\
Bottle placement, dual-arm & 0.517 & 0.250 \\
Bottle placement, single-arm & 1.040 & 0.664 \\
\midrule
Average & 0.884 & \textbf{0.520} \\
\bottomrule
\end{tabular}
\end{table}

With foundation pretraining, average next-event MSE falls from $0.884$ to $0.520$, a relative reduction of $41.2\%$, with improvements on all seven tasks. The largest absolute reduction occurs in tape-roll pick-and-place ($2.313$ to $1.634$), while cloth folding improves from $1.226$ to $0.503$. Precision insertion reaches the lowest error with pretraining ($0.170$), consistent with the clear pick-up, insertion, refinement, and completion boundaries in Figure~\ref{fig:precise_insert}. These results show that event-scale prediction transfers from the foundation checkpoint; the gate separately attributes the benefit of introducing the objective during pretraining.

\subsection{Uncertainty Identifies Hesitation and Failure}
\label{sec:uncertainty_analysis}
For action-error detection, we average propagated variance over the same executed horizon used to compute normalized squared action error. High-error chunks exceed the 90th-percentile error threshold selected on a disjoint validation split; AUPRC measures how well the uncertainty level ranks these sparse positives. Early warning instead uses the finite-difference uncertainty gradient $G_t=U_t-U_{t-1}$, which responds to abrupt increases rather than a high but stable uncertainty level. An alarm requires three consecutive gradient-threshold crossings. The threshold is selected on successful validation trajectories to give a 10\% trajectory-level false-alarm rate, and lead time is measured from the first sustained gradient alarm to annotated failure onset. Table~\ref{tab:uncertainty_pretrain_compare} evaluates uncertainty transfer from foundation pretraining against learning the complete architecture only from post-training data.

\begin{table}[H]
\centering
\caption{Action-uncertainty quality across seven real-world tasks. Both variants use the same complete architecture and post-training recipe; they differ only in whether training starts from the jointly pretrained checkpoint. AUPRC uses the uncertainty level; lead time uses uncertainty-gradient alarms set to a 10\% successful-trajectory false-alarm rate.}
\label{tab:uncertainty_pretrain_compare}
\scriptsize
\setlength{\tabcolsep}{2.7pt}
\resizebox{\linewidth}{!}{%
\begin{tabular}{@{}lrrrr@{}}
\toprule
& \multicolumn{2}{c}{Action-error AUPRC$\uparrow$} & \multicolumn{2}{c}{Failure lead time (s)$\uparrow$} \\
\cmidrule(lr){2-3}\cmidrule(lr){4-5}
Task & \shortstack{w/o\\pretraining} & \shortstack{w/\\pretraining} & \shortstack{w/o\\pretraining} & \shortstack{w/\\pretraining} \\
\midrule
Pick and place tape roll & 0.50& 0.53& 0.82& 1.85\\
Precision part insertion & 0.53& 0.68& 0.63& 1.39\\
Tableware organization & 0.49& 0.64& 1.13& 1.66\\
Cloth folding & 0.40& 0.47& 0.95& 1.44\\
Battery insertion & 0.46& 0.53& 0.94& 1.47\\
Bottle placement, dual-arm & 0.57& 0.66& 0.96& 1.45\\
Bottle placement, single-arm & 0.51& 0.63& 1.11& 1.54\\
\midrule
Average & 0.49 & \textbf{0.59} & 0.93 & \textbf{1.54} \\
\bottomrule
\end{tabular}%
}
\end{table}

With foundation pretraining, macro action-error AUPRC rises from $0.49$ to $0.59$, and mean gradient-based failure lead time increases from $0.93$\,s to $1.54$\,s ($+0.61$\,s) at the fixed 10\% false-alarm operating point. Both metrics improve on all seven tasks. The matched architecture and post-training recipe show that useful uncertainty transfers from the foundation checkpoint; component attribution is reported separately at gate scale.

We next connect the aggregate measurements to a recognizable contact-rich execution. We average propagated variance over the action dimensions used by the precision-insertion policy to obtain $U_t$, and use its gradient $G_t$ only for warning decisions.

\begin{figure}[t]
    \centering
    \includegraphics[width=\linewidth]{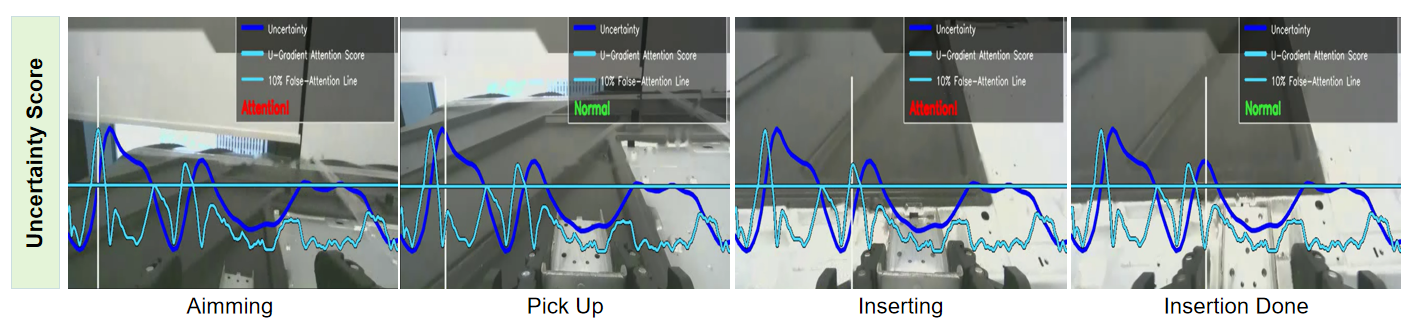}
    \caption{Uncertainty and uncertainty-gradient warning during precision part insertion. Blue shows the uncertainty level $U_t$, cyan shows its temporal gradient $G_t$, and the horizontal cyan line is the gradient threshold calibrated to a 10\% false-alarm rate on successful validation trajectories. White cursors mark the illustrated timesteps; red and green labels denote warning and normal states, respectively.}
    \label{fig:uncertainty_insert}
\end{figure}

Figure~\ref{fig:uncertainty_insert} separates uncertainty magnitude from the onset of uncertainty. During initial aiming, $U_t$ rises rapidly and $G_t$ crosses the warning threshold. Once the grasp is established, the gradient falls below threshold and the state returns to normal even though the uncertainty level remains temporarily elevated. A second gradient crossing occurs as the controller commits to insertion, where small pose errors can produce large action residuals. After the tip is seated, the uncertainty stabilizes and the warning clears. The gradient detector therefore responds to rapid loss of local reliability rather than treating every sustained uncertainty peak as a new alarm.

This pattern complements the RTG trace in Figure~\ref{fig:rtg_hard}. RTG accumulates task-scale progress through successful state transitions, whereas the uncertainty level measures local action reliability and its gradient detects abrupt deterioration within those transitions. Neither quantity is interpreted as a direct failure probability. Table~\ref{tab:uncertainty_pretrain_compare} evaluates error ranking from $U_t$ and warning time from $G_t$ at the 10\% false-alarm operating point. Additional success--failure and hesitation traces are provided in Appendix~\ref{app:uncertainty_examples}.

\section{Related Work}

\paragraph{Generalist vision-language-action policies.}
RT-1 and RT-2 established scalable transformer-based robot control and transfer from vision--language pretraining \citep{brohan2022rt1,brohan2023rt2}. Open X-Embodiment and Octo broadened cross-robot data and policy reuse \citep{openx2023,octo2024}; OpenVLA enabled open model adaptation \citep{kim2024openvla}; and $\pi_0$ introduced flow matching for a generalist VLA action expert \citep{black2024pi0}. GR00T N1 couples a vision--language module to a generative action module for cross-embodiment control \citep{bjorck2025gr00tn1}. These systems provide strong generalist backbones but primarily expose actions at inference time. \method{} extends the foundation-model interface itself: progress, event intention, and action reliability are jointly pretrained with action generation and remain available online.

\paragraph{Brain-inspired sensorimotor organization.}
The brain must control behavior whose relevant causes evolve at different rates. Temporal-hierarchy accounts propose that slowly evolving contextual states organize faster trajectories \citep{kiebel2008timescales,hasson2008hierarchy,murray2014intrinsic,chaudhuri2015hierarchical}; work on hierarchical behavior describes tasks as compositions of subtasks and primitive actions \citep{botvinick2009hierarchical}; and event-segmentation research links continuous experience to nested, behaviorally meaningful boundaries \citep{zacks2007event,baldassano2017event}. Outcome prediction, internal forward models, and probabilistic sensorimotor inference further connect action to expected consequences and reliability \citep{schultz1997reward,kording2004bayesian,wolpert1995sensorimotor,wolpert1998internal,knill2004bayesian,faisal2008noise}. These findings do not prescribe a robot architecture, but jointly motivate a functional prior: control should maintain predictive state at multiple timescales. \method{} operationalizes that prior as supervised targets and directed conditioning, not as a biological replica.

\paragraph{Value and progress estimation.}
Value functions estimate expected future return, but large imitation datasets rarely provide dense, comparable rewards, and expert demonstrations underrepresent low-quality states. We therefore construct an \emph{empirical} RTG target from observed terminal outcomes and remaining duration. Sequence models such as Decision Transformer use RTG to condition policies in offline reinforcement learning \citep{chen2021decision}; in the VLA setting, ReCAP for $\pi^{*}_{0.6}$ \citep{intelligence2025pi} uses a related ordering signal to rank nominal states by proximity to completion and separate failed endpoints with a terminal penalty. Our target is neither an optimal value function nor a safety certificate; it is a supervised progress-and-quality signal defined on both successful and failed trajectories. More recently, STEAM \citep{liu2026steam} treats temporally reversed episodes as failure samples, while PRTS \citep{zhang2026prts} introduces contrastive RL and trajectories paired with incorrect prompts as negative cases. In contrast, we use genuine policy failures and learn RTG during pretraining to support progress estimation on off-nominal states.

\paragraph{Hierarchical and subgoal-conditioned manipulation.}
Temporal abstraction is classically formalized through skills and options \citep{sutton1999options}. In modern robot learning, SayCan grounds language-model plans with skill values \citep{ahn2022saycan}, whereas RT-H predicts language motions before low-level actions \citep{belkhale2024rth}. Among VLA models, COT-VLA predicts keyframe images as visual chain-of-thought supervision before action prediction \citep{zhao2025cot}. $\pi_{0.5}$ defines events with textual prompts and co-trains textual subtasks and actions \citep{intelligence2025pi1}, while $\pi_{0.7}$ uses a smaller model to segment textual events and a world model to predict keyframe images as VLA priors \citep{intelligence2026pi,torne2026mem}. Because image- or context-level prediction can be computationally expensive, other work represents stages or keyframes through downsampled targets \citep{xu2026improving} or compact discrete spaces \citep{yang2026eventvla}. These hierarchies improve long-horizon structure, but often require an auxiliary generator, a skill library, or an additional language-to-control interface. ETG instead represents a subgoal as executable motion in the native action space and terminates it at a verified event. It therefore shares embodiment normalization and dynamics with the short-horizon action expert.


\paragraph{Uncertainty in generative policies and VLAs.}
Uncertainty estimators for generative models use conditional perturbations \citep{berry2024shedding}, Bayesian or Laplace approximations \citep{daxberger2021laplace,kou2024bayesdiff}, pixel-wise aleatoric uncertainty \citep{de2025diffusion}, or feature-space likelihoods \citep{radford2021learning,jazbec2025generative}. UA-Flow predicts velocity together with heteroscedastic uncertainty and propagates it through flow dynamics, but validates the formulation on image generation rather than robot control \citep{han2026flow}. VLA-specific uncertainty is more recent. EveryDayVLA uses disagreement between discrete and continuous action heads to adapt the execution horizon \citep{chopra2025everydayvla}; SCALE constructs self-uncertainty at inference time \citep{choi2026scale}; Tang et al. aggregate token entropy with motion-aware calibration \citep{tang2026shifting}; and R\"omer et al. use velocity-field disagreement across a small ensemble for failure detection and active fine-tuning \citep{romer2026uqvla}. These studies establish the value of VLA uncertainty, but derive it from output disagreement, token statistics, inference-time procedures, or multiple policies. \method{} instead learns heteroscedastic velocity variance as a native projection of the action expert during foundation pretraining and propagates it into action space without a detached verifier or policy ensemble.

\section{Conclusion}
Generalist VLAs have become increasingly capable, but their action-only interfaces leave progress, intermediate intention, and local reliability implicit. Inspired by functionally distinct predictive state in biological sensorimotor control, \method{} brings these variables into a pretrained, control-aligned explanatory interface. RTG reports \emph{whether} the state is progressing, ETG reports \emph{what} event-level transition is being pursued, and propagated flow variance reports \emph{how reliably} the motor command is generated. To our knowledge, \method{} is the first large-scale generalist VLA to learn heteroscedastic uncertainty natively inside its action expert. All three targets are trained jointly with action generation on a large mixture of successful and failed robot trajectories, emitted online, and kept inside the control pathway rather than attached after training.

The controlled attribution gate improves mean success by 16.0 percentage points over the action-only backbone, 10.8 points over the strongest single-head variant, and 6.8 points over an action-pretrained model that learns the same three heads only downstream. This evidence was obtained before, and used to authorize, the costly full run. After pretraining, \method{} achieves 74.1\% mean success across 50 randomized-hard RoboTwin2.0 tasks versus 55.4\% for GR00T N1.7, and improves mean real-robot success from 50.7\% to 73.5\%. Relative to learning the same architecture only from post-training data, foundation pretraining raises average RTG Spearman correlation from $0.71$ to $0.91$, reduces RTG MSE from $0.100$ to $0.050$, and reduces next-event ETG MSE from $0.884$ to $0.520$. It also raises action-error AUPRC from $0.49$ to $0.59$ and extends gradient-based failure lead time from $0.93$\,s to $1.54$\,s at a 10\% false-alarm rate. The temporal analyses show RTG responding to subgoal completion, target switching, and ambiguous grasps, while uncertainty rises during hesitation and its gradient marks rapid deterioration in local reliability. Together, these results support multi-timescale predictive state as a practical bridge between stronger control and intrinsic explainability. Counterfactual faithfulness tests and closed-loop recovery remain important next steps.

{\small
\bibliographystyle{plainnat}
\bibliography{references}
}
\newpage
\appendix
\section{Discussion}
\subsection{Why Task-Dependent Events Share One Interface}
Although the evidence used to locate events differs across task families, the learned representation is uniform: every event is an action chunk ending at a verified semantic boundary. The policy need not infer the annotation heuristic itself. At inference time, it receives only standard observations, language, and robot state; sensor thresholds, operator inputs, and manually specified poses are used exclusively to construct supervision.

The shared representation also supports failure attribution. In insertion, an incorrect pre-insertion event suggests an alignment error; in wiping, a correct stroke target followed by poor surface contact suggests an execution error; and in cloth manipulation, ambiguity near a stabilization boundary may be invisible from gripper state alone. ETG thus preserves task-specific semantics without introducing task-specific prediction heads.
\subsection{Why Multi-Timescale Prediction Can Improve Control}
The quantitative results indicate that the predictive variables act as more than a reporting interface. Joint pretraining gives the shared representation three operational targets that an action-only policy is not required to separate: RTG supplies dense episode-level ordering, ETG identifies the next behaviorally meaningful transition, and action flow resolves that transition while representing conditional residual scale. Because these variables are optimized with the backbone and condition the action pathway, they can shape control rather than merely describe a frozen policy after the fact. Failed trajectories are especially informative because they expose states in which nominal task phase, event intention, and local execution cease to agree. Although we do not directly measure the resulting representation geometry, the controlled gate places this interpretation on firmer ground: \method{} exceeds the strongest single-head variant by 10.8 points and the action-pretrained post-hoc-head control by 6.8 points. These gate-scale comparisons, rather than the later end-system comparison with GR00T, provided the component-level evidence for retaining the joint design before full pretraining.

These objectives may also discourage shortcuts: action-only imitation can map a scene to a command without representing whether it advances the task. RTG distinguishes progress from regression, ETG preserves the next semantic boundary, and heteroscedastic flow penalizes uniform confidence across variable residuals. This extra structure does not guarantee causal reasoning.

\subsection{What ``Brain-Inspired'' and ``Explainable'' Mean Here}

Calling a policy brain-inspired or explainable is useful only if those labels impose design commitments that could fail empirically. Here the commitments are: (i) control state is separated by temporal scale; (ii) each scale answers an explicit control question through direct supervision; (iii) slower predictions condition faster control, while variance is estimated inside the action expert; and (iv) all signals are emitted online with expected temporal signatures. Table~\ref{tab:brain_mapping} makes these commitments concrete. Before expensive pretraining, the pretraining-gate ablation tests whether each scale contributes enough to control to justify inclusion in the full run. The later held-out evaluation tests RTG progress ordering, ETG next-event prediction, the error alignment of action variance, and the early-warning behavior of its temporal gradient; visualizations then connect these aggregate scores to recognizable execution events.

\begin{table}[h]
\centering
\small
\caption{The explicit explanatory interface. Each signal answers a control-relevant question through a supervised target and testable temporal signature.}
\label{tab:brain_mapping}
\begin{tabular}{@{}p{0.27\linewidth}p{0.16\linewidth}p{0.48\linewidth}@{}}
\toprule
Explanatory question & Signal & Operational meaning and signature \\
\midrule
Is the state improving? & RTG & Task-scale progress; decreases under visible regression \\
What transition comes next? & ETG & Action chunk ending at the next verified task event \\
How reliable is local control? & Variance & Conditional residual scale; rises during hesitation or instability \\
\bottomrule
\end{tabular}
\end{table}

The claim remains functional, not mechanistic. RTG is not asserted to reproduce dopaminergic activity, ETG is not a model of cortical event-boundary detection, and heteroscedastic action flow is not a neural population code. Nor do our results establish that biological intelligence uses the same objectives. The brain literature motivates the computational problem---prediction under nested environmental dynamics---and \method{} provides one engineering realization whose consequences can be tested. Anatomical localization and biological plausibility remain outside the scope of this work.

\subsection{Scope of Explainability}
\method{} provides intrinsic predictive signals rather than natural-language rationales or formal guarantees. RTG, ETG, and variance are \emph{explicit by construction}: their supervised targets correspond to progress, semantic events, and action dispersion, and their predictions are available alongside every action. They are also more control-aligned than a detached explanation head because RTG and ETG condition action features and variance shares the action expert. This supports process-level inspection---what the policy believes about state, intent, and reliability---rather than only outcome-level inspection of the executed command.

Explicitness does not guarantee causal faithfulness. The model may exploit correlates such as hand visibility, object appearance, camera motion, or episode timing. Our RTG examples show non-monotonic behavior inconsistent with elapsed time alone; counterfactual interventions are still needed to identify which visual cues cause each prediction. Likewise, a well-formed ETG can expose the policy's predicted intention without proving that this intention is causally decisive for the sampled action.

ETG is also bounded by the event vocabulary. Unlabeled transitions cannot be exposed explicitly, while overly dense annotations collapse toward ordinary action chunking and lose temporal abstraction. Our task-dependent procedure balances these extremes by preserving manipulation-specific semantics within a shared action representation.

\subsection{Deployment Implications}
The outputs naturally support online monitoring. An interface can display the predicted event, normalized progress, arm-specific uncertainty, and uncertainty-gradient warning state while logging each signal alongside video for failure analysis. More actively, a gradient alarm could trigger slower control or recovery, RTG could detect stalled execution, and ETG could provide a replanning checkpoint. Because these quantities are produced by the policy itself, they are available at every inference step without external oracle modules.

Deployment requires safeguards against false confidence. Low variance is meaningful only relative to the training distribution and does not certify that ETG is correct; similarly, RTG may fall because of occlusion rather than physical regression. Safety-critical systems should combine these signals with state constraints, contact monitoring, collision checking, and independent anomaly detection. \method{} increases policy visibility but does not replace system-level safety mechanisms.

\subsection{Limitations and Future Evaluation}
Our signal-level study quantifies progress ranking and error, next-event action error, uncertainty precision--recall, and gradient-based failure-warning lead time across the seven real-world tasks. The warning threshold is set to a 10\% false-alarm rate on successful validation trajectories, but these metrics do not by themselves establish causal faithfulness or guarantee transfer to unseen tasks and embodiments. Controlled perturbations---including occlusion, object displacement, unexpected contact, and instruction changes---would further test whether each signal responds to known causal factors. ETG boundary timing, event-horizon coverage, and semantic consistency under novel event compositions also remain useful complementary evaluations.

The real-world evaluation is limited to 20 trials per task, one checkpoint per method, and no confidence intervals. Likewise, the simulation study uses a single training run per configuration. Repeated seeds and per-task uncertainty estimates are needed to quantify statistical robustness. The controlled gate includes the matched action-pretrained post-hoc-head comparison needed for component attribution, but its five-task scale cannot guarantee that the same effect size holds over the full 20,000-hour mixture. Repeating the full run with an action-only foundation model would require a second pretraining campaign of comparable cost; accordingly, we attribute components at gate scale and treat the full-scale baseline comparisons as end-system evaluations. Adaptive loss weighting or task-conditioned routing may further address the task-specific brittleness exposed by Table~\ref{tab:ablation}. Finally, we evaluate the signals primarily as predictions; the next step is to use them to select recovery actions and measure their effect on safety and completion rate.

\section{Dataset Details}
The dataset covers heterogeneous embodiments, sensing configurations, and manipulation domains. Most platforms provide three to six camera streams and operate at 30--60 Hz; we downsample all trajectories to 30 Hz for training. Tasks retain their complete language instructions rather than short verb--noun templates. For example, we preserve ``Place the green and blue objects to the right of the pink plate using the right hand, and place the green and yellow objects to the left of the blue plate using the left hand,'' instead of reducing it to ``put the objects into the plates.'' Full instructions preserve object attributes, spatial relations, arm assignments, and ordering constraints that are essential for multi-object behavior.

\subsection{Embodiments}

\begin{figure*}[t]
    \centering
    \begin{subfigure}[t]{0.19\textwidth}
        \centering
        \includegraphics[width=\linewidth,height=3.1cm,keepaspectratio]
        {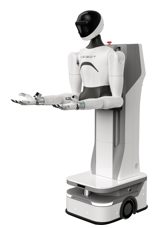}
        \caption{AgiBot G1}
        \label{fig:embodiment_agibot_g1}
    \end{subfigure}\hfill
    \begin{subfigure}[t]{0.19\textwidth}
        \centering
        \includegraphics[width=\linewidth,height=3.1cm,keepaspectratio]
        {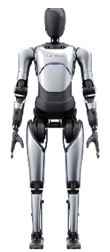}
        \caption{AgiBot A2}
        \label{fig:embodiment_agibot_a2}
    \end{subfigure}\hfill
    \begin{subfigure}[t]{0.19\textwidth}
        \centering
        \includegraphics[width=\linewidth,height=3.1cm,keepaspectratio]
        {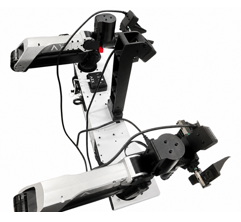}
        \caption{AC ONE}
        \label{fig:embodiment_ac_one}
    \end{subfigure}\hfill
    \begin{subfigure}[t]{0.19\textwidth}
        \centering
        \includegraphics[width=\linewidth,height=3.1cm,keepaspectratio]
        {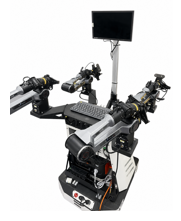}
        \caption{AgileX-Aloha}
        \label{fig:embodiment_cobot_magic}
    \end{subfigure}\hfill
    \begin{subfigure}[t]{0.19\textwidth}
        \centering
        \includegraphics[width=\linewidth,height=3.1cm,keepaspectratio]
        {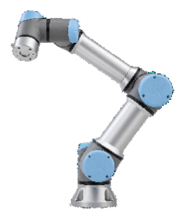}
        \caption{UR5e}
        \label{fig:embodiment_ur5e}
    \end{subfigure}

    \vspace{0.6em}

    \begin{subfigure}[t]{0.19\textwidth}
        \centering
        \includegraphics[width=\linewidth,height=3.1cm,keepaspectratio]
        {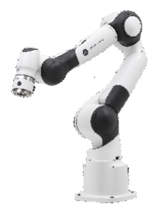}
        \caption{Franka}
        \label{fig:embodiment_franka}
    \end{subfigure}\hfill
    \begin{subfigure}[t]{0.19\textwidth}
        \centering
        \includegraphics[width=\linewidth,height=3.1cm,keepaspectratio]
        {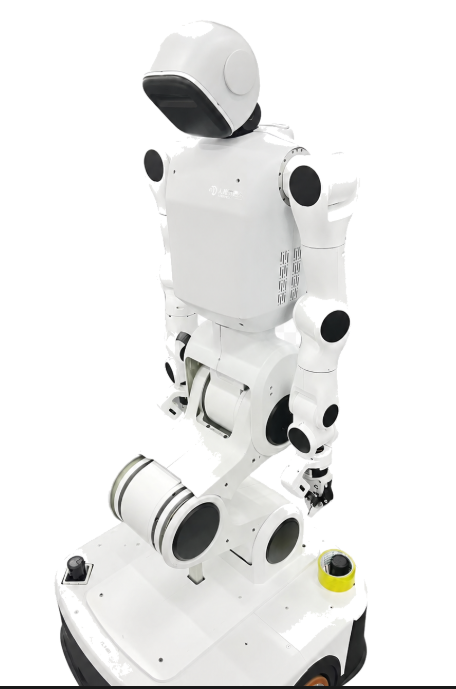}
        \caption{Loong S1}
        \label{fig:embodiment_qloong_s1}
    \end{subfigure}\hfill
    \begin{subfigure}[t]{0.19\textwidth}
        \centering
        \includegraphics[width=\linewidth,height=3.1cm,keepaspectratio]
        {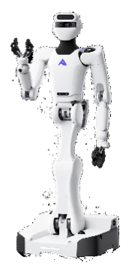}
        \caption{Astribot S1}
        \label{fig:embodiment_astribot_s1}
    \end{subfigure}\hfill
    \begin{subfigure}[t]{0.19\textwidth}
        \centering
        \includegraphics[width=\linewidth,height=3.1cm,keepaspectratio]
        {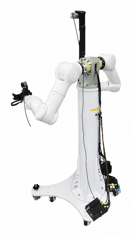}
        \caption{Tianji Marvin}
        \label{fig:embodiment_tianji_marvin}
    \end{subfigure}\hfill
    \begin{subfigure}[t]{0.19\textwidth}
        \centering
        \includegraphics[width=\linewidth,height=3.1cm,keepaspectratio]
        {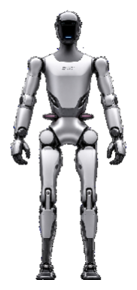}
        \caption{Fourier GR2}
        \label{fig:embodiment_fourier_gr2}
    \end{subfigure}

    \caption{Representative robot embodiments in our dataset.}
    \label{fig:robot_embodiments}
\end{figure*}

Most embodiments use 1-DoF parallel-jaw grippers; selected AgiBot A2, Tienkung, and Fourier configurations instead use 6-DoF dexterous hands. Figure~\ref{fig:robot_embodiments} illustrates 10 of them.
\paragraph{Bimanual UR5e.} This platform comprises two UR5e arms with parallel-jaw grippers, two wrist cameras, and one over-the-shoulder camera. Its configuration and action spaces are 14-dimensional.
\paragraph{Bimanual Franka.} This platform comprises two Franka arms with parallel-jaw grippers and wrist-mounted cameras. An over-the-shoulder view and, in some configurations, an additional third-person camera yield three or four image streams. Its configuration and action spaces are 16-dimensional.
\paragraph{Bimanual ARX (AC One).} This fixed-base platform uses two 6-DoF ARX arms, two wrist cameras, and one base camera, with 14-dimensional configuration and action spaces.
\paragraph{Bimanual ARX (Mobile).} This embodiment uses the same dual 6-DoF ARX arms, three-camera layout, and 14-dimensional spaces as AC One, but mounts the system on an AgileX mobile base with 3 DoF.
\paragraph{AgileX-Aloha (Cobot Magic).} This fixed-base platform uses two 6-DoF arms, two wrist cameras, and one base camera. Its configuration and action spaces are 14-dimensional; its kinematic structure differs from the bimanual ARX platform at two joints.

\paragraph{AgiBot G1.} This humanoid has 14 arm DoF (7 per arm), one lift DoF, and two gripper DoF, together with two hand cameras and one head camera.
\paragraph{AgiBot A2.} This humanoid has 14 arm DoF, two head DoF, and 12 hand DoF (6 per hand), with two chest cameras and one head camera.
\paragraph{Fourier GR2.} This humanoid has 41 DoF spanning the shoulders, hands, waist, wrists, knees, and ankles, and uses two head cameras.
\paragraph{Galaxea R1.} This humanoid has 12 arm DoF (6 per arm), three lift DoF, and two gripper DoF, with two hand cameras and one head camera.
\paragraph{Qloong 1 and Qloong 2.} Each humanoid has 14 arm DoF, one lift DoF, two gripper DoF, two hand cameras, and one head camera. Both platforms are produced by Humanoid Robot (Shanghai) Co., Ltd.
\paragraph{Leju KUAVO.} This humanoid has 14 arm DoF, two head DoF, 12 leg DoF, and two gripper DoF, with two hand cameras and one head camera.
\paragraph{Astribot S1.} This humanoid has 25 DoF spanning the arms, shoulders, head, and hips, together with two hand cameras, one head camera, and one torso camera.
\paragraph{Tienkung.} This humanoid has 14 arm DoF and 12 hand DoF, with one head camera. Some configurations replace the dexterous hands with 2-DoF grippers.
\paragraph{Tianji Marvin.} This humanoid has 14 arm DoF and two gripper DoF, with two hand cameras and one head camera.
\paragraph{Dwheel.} This mobile variant mounts Tianji Marvin on a wheeled base, adding two wheel DoF to its 14 arm and two gripper DoF. It uses two hand cameras and one head camera.
\paragraph{Loong S1.} This wheeled mobile manipulator has two 7-DoF arms, two gripper DoF, two wrist cameras, and one head camera. It was released by Humanoid Robot (Shanghai) Co., Ltd. in August 2026. No Loong S1 trajectory appears in the pretraining mixture, so the two real-world tasks on this platform evaluate transfer to an unseen embodiment after task-specific post-training.

\begin{figure}[t]
    \centering
    \includegraphics[width=\linewidth]{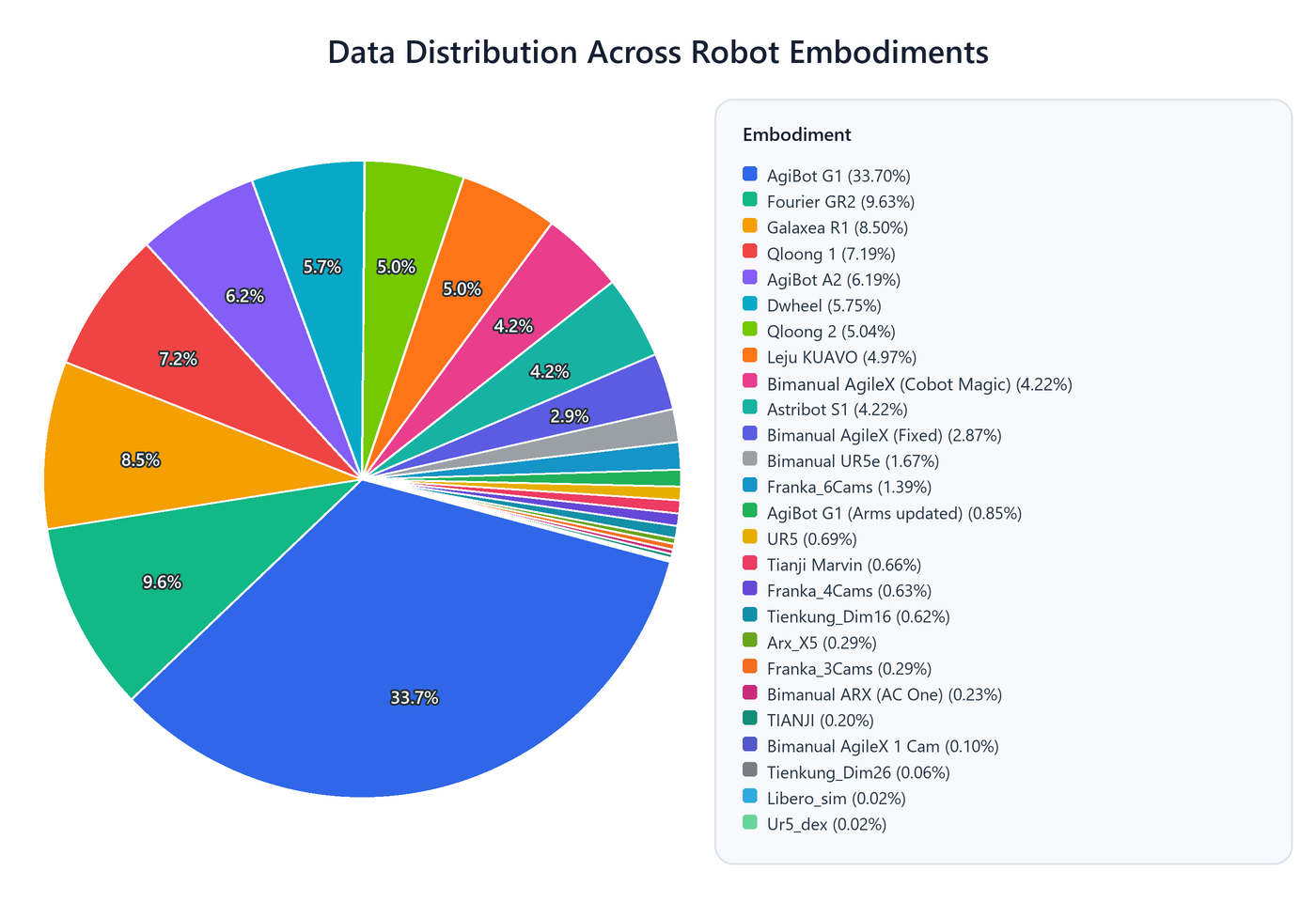}
    \caption{Data distribution across robot embodiments.}
    \label{fig:data_r}
\end{figure}

The embodiment distribution is summarized in Table~\ref{tab:data_ratio} and Figure~\ref{fig:data_r}. We additionally include the open-source OXE \citep{o2024open} and AgiBot World \citep{bu2025agibot} datasets, sampling from this public-data pool with a probability of 10\%.

\begin{table}[t]
\centering
\small
\caption{Data distribution across robot embodiments.}
\label{tab:data_ratio}
\begin{tabular}{lr}
\toprule
Embodiment & Ratio \\
\midrule
AgiBot G1 & 33.70\% \\
Fourier GR2 & 9.63\% \\
Galaxea R1 & 8.50\% \\
Qloong 1 & 7.19\% \\
AgiBot A2 & 6.19\% \\
Dwheel & 5.75\% \\
Qloong 2 & 5.04\% \\
Leju KUAVO & 4.97\% \\
Bimanual AgileX (Cobot Magic) & 4.22\% \\
Astribot S1 & 4.22\% \\
Bimanual AgileX (Fixed) & 2.87\% \\
Bimanual UR5e & 1.67\% \\
Franka\_6Cams & 1.39\% \\
AgiBot G1 (Arms updated) & 0.85\% \\
UR5 & 0.69\% \\
Tianji Marvin & 0.66\% \\
Franka\_4Cams & 0.63\% \\
Tienkung\_Dim16 & 0.62\% \\
Arx\_X5 & 0.29\% \\
Franka\_3Cams & 0.29\% \\
Bimanual ARX (AC One) & 0.23\% \\
TIANJI & 0.20\% \\
Bimanual AgileX 1 Cam & 0.10\% \\
Tienkung\_Dim26 & 0.06\% \\
Libero\_sim & 0.02\% \\
Ur5\_dex & 0.02\% \\
\bottomrule
\end{tabular}
\end{table}

\subsection{Event Definition}\label{app:event_define}
\paragraph{Pick and place.}
Pick-and-place tasks, including table arrangement and sorting, contain informative gripper transitions. Rising and falling edges in the gripper signal propose pre-grasp, grasp, transfer, and release events. Human verification removes transitions caused by incidental gripper motion and aligns each retained event with the corresponding visual state.

\paragraph{Insertion.}
Gripper transitions alone do not identify the critical pre-insertion event: the aligned pose immediately before the object enters the target. During VR teleoperation, the operator marks this pose with a designated input; annotators subsequently verify the marker against the recorded observations and motion.

\paragraph{Deformable-object manipulation.}
Cloth folding and stacking involve frequent gripper motion and geometrically ambiguous states. Near-zero joint velocity proposes stabilization boundaries, while task-specific poses add semantic structure. For garment manipulation, we annotate the highest lifting pose because it separates acquisition from hanging or folding. These tasks require more manual verification than rigid-object transfer.

\paragraph{Wiping.}
For table and whiteboard wiping, contact-circuit feedback identifies initial tool--surface contact, and turning points in the repeated trajectory define subsequent events. These boundaries mark completed strokes rather than gripper transitions, which may remain unchanged throughout the task.

\section{Experiment Details}
\subsection{Component-Attribution Gate Details}\label{app:ablation}
\begin{figure}[t]
    \centering

    \begin{subfigure}{\linewidth}
        \centering
        \includegraphics[width=\linewidth]{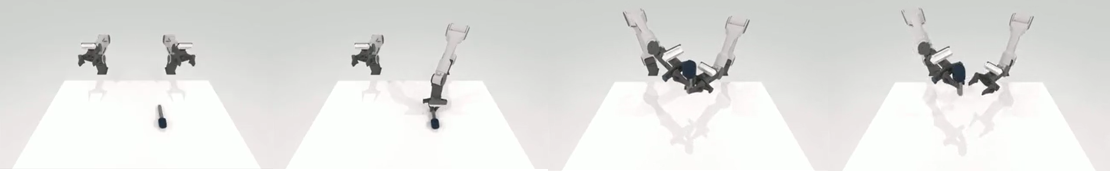}
        \caption{Handover microphone}
        \label{fig:handover_mic}
    \end{subfigure}

    \vspace{0.5em}

    \begin{subfigure}{\linewidth}
        \centering
        \includegraphics[width=\linewidth]{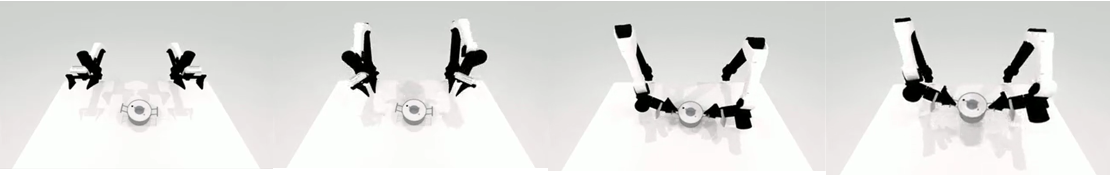}
        \caption{Lift pot}
        \label{fig:lift_pot}
    \end{subfigure}

    \vspace{0.5em}

    \begin{subfigure}{\linewidth}
        \centering
        \includegraphics[width=\linewidth]{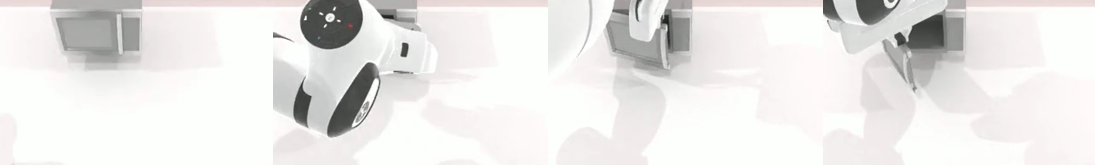}
        \caption{Open microwave}
        \label{fig:open_micro}
    \end{subfigure}

    \vspace{0.5em}

    \begin{subfigure}{\linewidth}
        \centering
        \includegraphics[width=\linewidth]{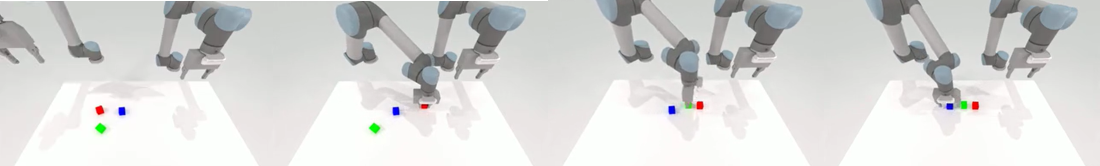}
        \caption{Rank blocks by color}
        \label{fig:ranking_rgb}
    \end{subfigure}

    \vspace{0.5em}

    \begin{subfigure}{\linewidth}
        \centering
        \includegraphics[width=\linewidth]{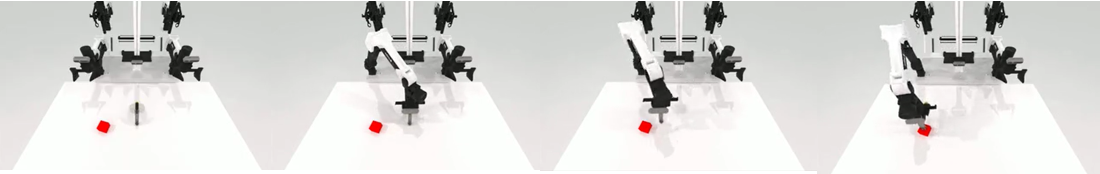}
        \caption{Hit block with hammer}
        \label{fig:beat_block}
    \end{subfigure}

    \caption{Execution processes of five representative manipulation tasks.}
    \label{fig:five_task_processes}
\end{figure}

As described in Section~\ref{sec:ablation}, this study is run before large-scale real-robot pretraining as a lower-cost component-attribution gate for the RTG, ETG, and uncertainty modules. Its purpose is to reveal ineffective components, harmful interactions, and whether the three objectives must participate during pretraining before committing approximately 20 days on 64 B200 GPUs. We exclude five of the 50 RoboTwin2.0 tasks from gate pretraining and reserve them for downstream post-training and evaluation. The OOD tasks are microphone handover, pot lifting, microwave opening, color-based block ranking, and hammering, each instantiated on different simulation embodiments: Piper, ARX, Franka, UR, or Aloha. Figure~\ref{fig:five_task_processes} shows representative executions, and Table~\ref{tab:ablation_setting} summarizes the gate protocol.

\begin{table}[t]
\centering
\caption{Held-out RoboTwin2.0 setup for the pretraining-gate ablation. All evaluation trials use randomized task configurations.}
\label{tab:ablation_setting}
\small
\renewcommand{\arraystretch}{1.08}
\setlength{\tabcolsep}{6pt}
\begin{tabular}{@{}p{0.40\linewidth}ccc@{}}
\toprule
& \multicolumn{2}{c}{Post-training episodes} & Test trials \\
\cmidrule(lr){2-3}\cmidrule(l){4-4}
Task & Clean & Randomized & Randomized \\
\midrule
Handover microphone & 50 & 200 & 50 \\
Lift pot & 50 & 200 & 50 \\
Open microwave & 50 & 200 & 50 \\
Rank blocks by color & 50 & 200 & 50 \\
Hit block with hammer & 50 & 200 & 50 \\
\bottomrule
\end{tabular}
\end{table}
In \emph{handover microphone}, the robot grasps a microphone with the nearer arm and transfers it to the other arm. In \emph{lift pot}, it grasps both handles and lifts the pot bimanually. In \emph{open microwave}, one arm pulls the handle and continues the motion until the door is open. In \emph{rank blocks by color}, the robot arranges red, green, and blue blocks from left to right. In \emph{hit block with hammer}, it grasps the hammer and strikes the target block.

All gate variants use the same 45-task pretraining mixture, number of optimization steps, observation and action representations, and downstream protocol. For the post-hoc control, we first optimize only the deterministic action objective on the 45-task mixture. We then instantiate the RTG, ETG, and uncertainty heads and jointly post-train the complete architecture on the five held-out tasks. The jointly pretrained LM-X variant has the same downstream architecture, annotations, and optimization, but includes all three explanatory objectives during 45-task pretraining. Consequently, the 72.8\% versus 79.6\% comparison in Table~\ref{tab:ablation} isolates, at gate scale, whether the explanatory objectives enter during pretraining or only during downstream adaptation.

For each task, we collect 20 failed episodes for LM-RTG and \method{}. We use these episodes only for RTG supervision: every 20 pretraining steps, a failure minibatch updates the RTG branch while gradients to the ETG and action experts are disabled. Post-training uses successful episodes only.

\subsection{Pretraining Recipe and Details}\label{app:train_recipe}
We begin full real-robot pretraining only after the controlled gate supports the complete three-module design against both the isolated objectives and the post-hoc-head control. We initialize \method{} from Cosmos-Reason2-2B \citep{agarwal2025cosmos} and optimize Eq.~\eqref{eq:global_loss} with $\lambda_R=0.1$, $\lambda_E=1$, and $\lambda_A=1$. Pretraining runs for two epochs on 64 NVIDIA B200 GPUs with a global batch size of 3,072, totaling approximately 700,000 gradient steps and finishing in about 20 days.

We sample failed episodes every 100 pretraining steps and use them only to update the RTG branch; the ETG and action experts are frozen for those steps.

Unless stated otherwise, all component variants share the architecture, observation preprocessing, action representation, and applicable loss definitions of the full model. During post-training, the model learns solely from successful episodes.

\subsection{Real-World Test Details}\label{app:real_setting}

For each real-world task, we fine-tune the complete pretrained model for two epochs on task-specific post-training data. VLA baselines are initialized from their official checkpoints and adapted with the same downstream data. All methods receive the same language instruction and are evaluated under the same task-specific success criterion.

\paragraph{Tape-roll pick-and-place.} The tape roll is initialized uniformly within an approximately 30 cm $\times$ 30 cm region. Success requires placing it inside a 20 cm $\times$ 20 cm target area marked with black tape.
\paragraph{Precision part insertion.} The robot must pick up the part, insert its 3 mm $\times$ 5 mm tip into the narrow slot, and refine the pose until two small screw tips enter their corresponding holes.
\paragraph{Tableware organization.} Six target objects are sampled from a set of 20 objects spanning five colors and randomly placed on the table. The robot must sort them into baskets, boxes, and plates according to color.
\paragraph{Cloth folding.} The robot must fold a yellow T-shirt initialized in a flattened configuration.
\paragraph{Battery insertion.} The robot must sequentially pick up four batteries and insert each into its designated slot.
\paragraph{Water-bottle placement, dual-arm.} The robot must grasp a standing bottle with the left hand, move the tray to a target position with the right hand, and place the bottle on the tray.
\paragraph{Water-bottle placement, single-arm.} The robot must use its right hand to grasp a standing bottle, reposition the tray, and place the bottle on it.

The reported comparisons use GR00T N1.7 \citep{nvidia2026gr00t17} and $\pi_{0.5}$ \citep{intelligence2025pi1}, each initialized from its released checkpoint and adapted on the same task-specific demonstration mixture with identical observations, instructions, success criteria, and evaluation trials. Model-specific optimization hyperparameters are selected on disjoint validation data.

\section{More Results}
\subsection{Ablation of Event Embeddings during Post-Training}\label{app:add_event}
Table~\ref{tab:add_event1} compares task-specific post-training with and without $\mathbf{z}^{\mathrm{event}}_t$ as an input to the action expert. During pretraining, the event embedding is retained with probability 80\% because of the 20\% dropout described in Section~\ref{sec:ablation}; here, we isolate whether event conditioning should remain active during post-training. It helps block ranking and hammering, is neutral for microphone handover, and hurts pot lifting and microwave opening. We therefore select this option per task using downstream validation performance. Event conditioning is enabled for block ranking and hammering in the gate experiments, and for precision insertion, tableware organization, and dual-arm water-bottle placement in the real-world experiments; it is disabled for the remaining tasks.
\begin{table}[t]
\centering
\small
\caption{Post-training ablation of event conditioning on five held-out RoboTwin2.0 tasks. Values are success rates (\%).}
\label{tab:add_event1}
\resizebox{\linewidth}{!}{%
\begin{tabular}{lrrrrr}
\toprule
Method &Handover microphone &Lift pot &Open microwave &Rank blocks &Hit block with hammer\\
\midrule
LM-X (with events) &\textbf{88} &80 &2 &\textbf{90} &\textbf{66}\\
LM-X (without events) &\textbf{88} &\textbf{92} &\textbf{62} &78 &60\\

\bottomrule
\end{tabular}
}
\end{table}
\subsection{RoboTwin2.0 Benchmark}\label{app:sim_result}
Table~\ref{tab:robotwin_step_comparison_full} reports the full 50-task RoboTwin2.0 results.
\begin{longtable}{lcc}
\caption{Performance comparison on RoboTwin2.0 tasks.}
\label{tab:robotwin_step_comparison_full} \\
\toprule
Task & GR00T N1.7 & \method{} \\
\midrule
\endfirsthead
\toprule
Task & GR00T N1.7 & \method{} \\
\midrule
\endhead
\texttt{adjust\_bottle} & 98.0 & 100.0 \\
\texttt{beat\_block\_hammer} & 40.0 & 25.0 \\
\texttt{blocks\_ranking\_rgb} & 60.0 & 93.0 \\
\texttt{blocks\_ranking\_size} & 52.0 & 82.0 \\
\texttt{click\_alarmclock} & 100.0 & 100.0 \\
\texttt{click\_bell} & 99.0 & 99.0 \\
\texttt{dump\_bin\_bigbin} & 77.0 & 85.0 \\
\texttt{grab\_roller} & 97.0 & 100.0 \\
\texttt{handover\_block} & 0.0 & 0.0 \\
\texttt{handover\_mic} & 69.0 & 87.0 \\
\texttt{hanging\_mug} & 0.0 & 33.0 \\
\texttt{lift\_pot} & 38.0 & 84.0 \\
\texttt{move\_can\_pot} & 72.0 & 86.0 \\
\texttt{move\_pillbottle\_pad} & 44.0 & 90.0 \\
\texttt{move\_playingcard\_away} & 48.0 & 19.0 \\
\texttt{move\_stapler\_pad} & 30.0 & 54.0 \\
\texttt{open\_laptop} & 74.0 & 96.0 \\
\texttt{open\_microwave} & 13.0 & 55.0 \\
\texttt{pick\_diverse\_bottles} & 63.0 & 85.0 \\
\texttt{pick\_dual\_bottles} & 79.0 & 96.0 \\
\texttt{place\_a2b\_left} & 39.0 & 50.0 \\
\texttt{place\_a2b\_right} & 25.0 & 46.0 \\
\texttt{place\_bread\_basket} & 76.0 & 91.0 \\
\texttt{place\_bread\_skillet} & 79.0 & 83.0 \\
\texttt{place\_burger\_fries} & 76.0 & 98.0 \\
\texttt{place\_can\_basket} & 33.0 & 35.0 \\
\texttt{place\_cans\_plasticbox} & 33.0 & 74.0 \\
\texttt{place\_container\_plate} & 97.0 & 97.0 \\
\texttt{place\_dual\_shoes} & 33.0 & 74.0 \\
\texttt{place\_empty\_cup} & 87.0 & 98.0 \\
\texttt{place\_fan} & 13.0 & 18.0 \\
\texttt{place\_mouse\_pad} & 34.0 & 53.0 \\
\texttt{place\_object\_basket} & 65.0 & 59.0 \\
\texttt{place\_object\_scale} & 41.0 & 58.0 \\
\texttt{place\_object\_stand} & 73.0 & 83.0 \\
\texttt{place\_phone\_stand} & 0.0 & 41.0 \\
\texttt{place\_shoe} & 85.0 & 96.0 \\
\texttt{press\_stapler} & 81.0 & 94.0 \\
\texttt{put\_bottles\_dustbin} & 1.0 & 64.0 \\
\texttt{put\_object\_cabinet} & 26.0 & 74.0 \\
\texttt{rotate\_qrcode} & 23.0 & 93.0 \\
\texttt{scan\_object} & 59.0 & 67.0 \\
\texttt{shake\_bottle} & 99.0 & 100.0 \\
\texttt{shake\_bottle\_horizontally} & 99.0 & 100.0 \\
\texttt{stack\_blocks\_three} & 33.0 & 68.0 \\
\texttt{stack\_blocks\_two} & 72.0 & 97.0 \\
\texttt{stack\_bowls\_three} & 63.0 & 75.0 \\
\texttt{stack\_bowls\_two} & 91.0 & 97.0 \\
\texttt{stamp\_seal} & 22.0 & 74.0 \\
\texttt{turn\_switch} & 61.0 & 79.0 \\
\midrule
\textbf{Average} & \textbf{55.4} & \textbf{74.1} \\
\bottomrule
\end{longtable}

\subsection{A Complementary RTG Case}\label{app:rtg_fail_case}
\begin{figure*}[t]
    \centering
    \begin{subfigure}[t]{0.24\textwidth}
        \includegraphics[width=\linewidth]{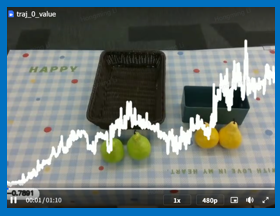}
        \caption{Start}
    \end{subfigure}
    \begin{subfigure}[t]{0.24\textwidth}
        \includegraphics[width=\linewidth]{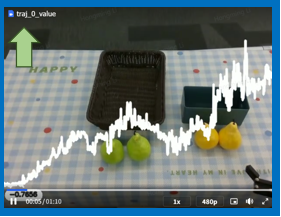}
        \caption{Right gripper appears}
    \end{subfigure}
    \begin{subfigure}[t]{0.24\textwidth}
        \includegraphics[width=\linewidth]{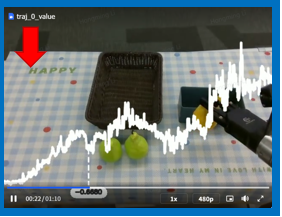}
        \caption{Stem grasp}
    \end{subfigure}
    \begin{subfigure}[t]{0.24\textwidth}
        \includegraphics[width=\linewidth]{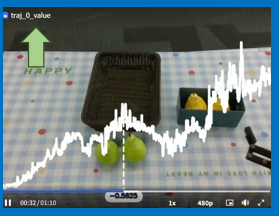}
        \caption{Halfway done}
    \end{subfigure}

    \vspace{0.5em}

    \begin{subfigure}[t]{0.24\textwidth}
        \includegraphics[width=\linewidth]{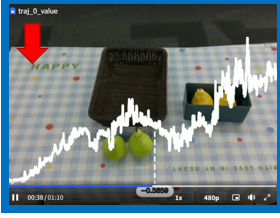}
        \caption{Gripper out of view}
    \end{subfigure}
    \begin{subfigure}[t]{0.24\textwidth}
        \includegraphics[width=\linewidth]{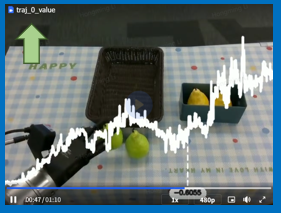}
        \caption{Left gripper appears}
    \end{subfigure}
    \begin{subfigure}[t]{0.24\textwidth}
        \includegraphics[width=\linewidth]{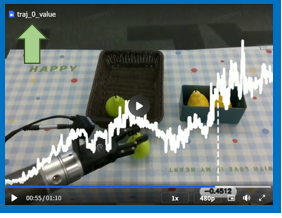}
        \caption{Stable progress}
    \end{subfigure}
    \begin{subfigure}[t]{0.24\textwidth}
        \includegraphics[width=\linewidth]{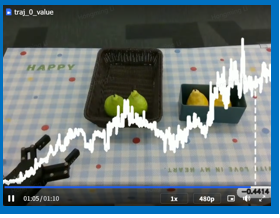}
        \caption{Finish}
    \end{subfigure}

    \caption{RTG visualization during a held-out pear-sorting episode. The score generally increases with progress but decreases at (c), where the right gripper holds only the pear's stem. Subsequent correctly executed subgoals restore the upward trend.}
    \label{fig:rtg_pears}
\end{figure*}
Figure~\ref{fig:rtg_pears} illustrates a complementary case. The policy must sort yellow and green pears into different containers. RTG rises as the right gripper enters view and begins the first subgoal, but decreases in panel (c) after the pear is lifted by its thin stem rather than its body. From the available views, this grasp appears ambiguous and potentially unstable.

No comparable decrease occurs for pears grasped around their bodies, suggesting that the score captures more than gripper closure or object elevation. Correct execution of later subgoals restores the upward trend, showing that a local decrease need not dominate the remainder of the trajectory.

This example also exposes a limitation: a vision-only progress head may conflate perceptual ambiguity with physical instability. If the stem grasp is mechanically secure but visually occluded, the lower RTG reflects uncertainty in the observation rather than true regression. Multi-view sensing and contact feedback may help disambiguate these cases. We therefore interpret the result as behaviorally responsive state evaluation, not ground-truth value estimation.

\subsection{Additional Uncertainty Visualizations}
\label{app:uncertainty_examples}
These extended traces complement the precision-insertion analysis in Section~\ref{sec:uncertainty_analysis} with held-out RoboTwin2.0 rollouts and a longer bimanual manipulation sequence.

\begin{figure*}[t]
    \centering
    \begin{subfigure}[t]{0.48\textwidth}
        \centering
        \includegraphics[width=\linewidth]{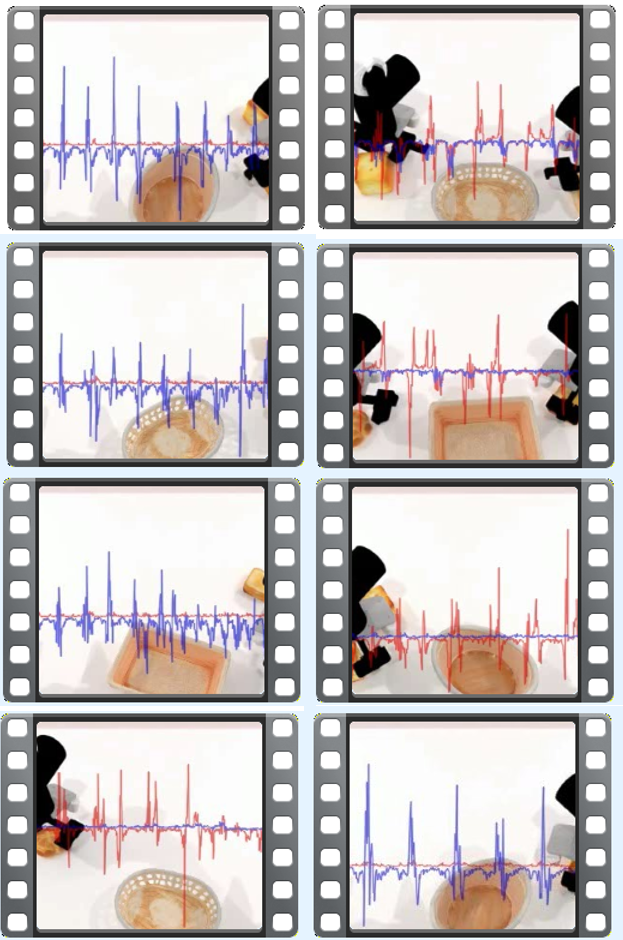}
        \caption{Failure cases.}
        \label{fig:failure_cases}
    \end{subfigure}\hfill
    \begin{subfigure}[t]{0.48\textwidth}
        \centering
        \includegraphics[width=\linewidth]{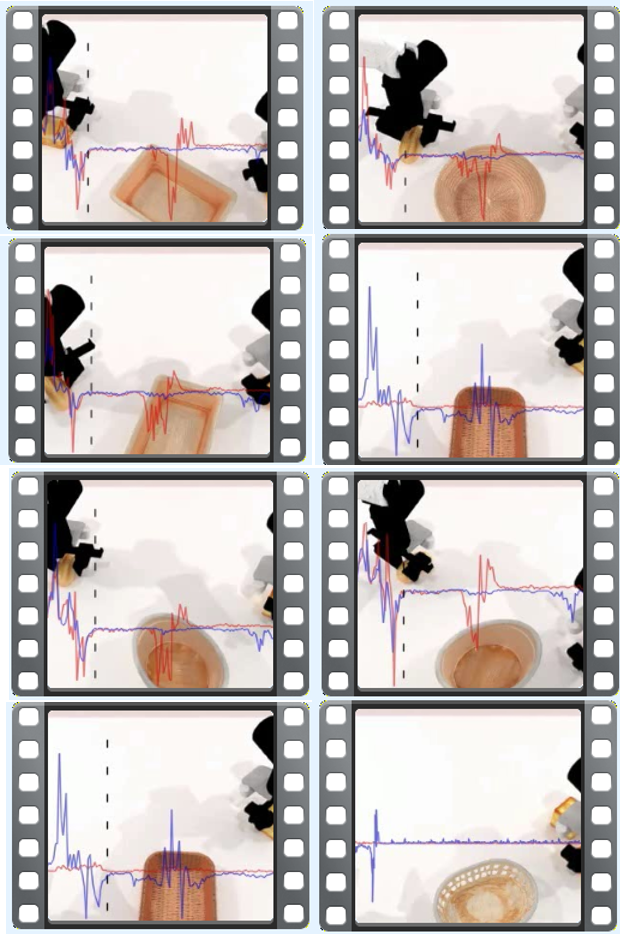}
        \caption{Success cases.}
        \label{fig:success_cases}
    \end{subfigure}
    \caption{Uncertainty traces for failed (left) and successful (right) held-out RoboTwin2.0 episodes. Blue and red curves average selected action variances for the right and left arms, respectively. Failed episodes show more frequent and larger spikes, whereas successful episodes remain comparatively stable after initial transients.}
    \label{fig:uncertainty_success_failure}
\end{figure*}

Figure~\ref{fig:uncertainty_success_failure} compares successful and failed rollouts. Failed episodes exhibit repeated variance spikes in one or both arms, whereas successful episodes remain lower and smoother after the initial transient. Failure is therefore associated not only with a terminal peak but with recurring intervals of ambiguous local control.

Arm-specific traces provide additional diagnostic structure: unilateral peaks suggest a local grasp or placement issue, whereas synchronized peaks may indicate a coordination problem. A single scalar confidence score would obscure this distinction. Although the figure does not establish a calibrated threshold, it shows that variance can be localized to subsets of the action space.

\begin{figure*}[t]
    \centering
    \begin{subfigure}[t]{0.24\textwidth}
        \includegraphics[width=\linewidth]{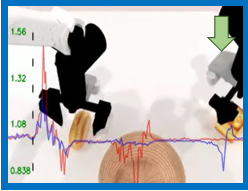}
        \caption{Aiming well}
    \end{subfigure}
    \begin{subfigure}[t]{0.24\textwidth}
        \includegraphics[width=\linewidth]{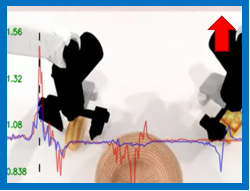}
        \caption{Hesitating while adjusting the distance to the table surface}
    \end{subfigure}
    \begin{subfigure}[t]{0.24\textwidth}
        \includegraphics[width=\linewidth]{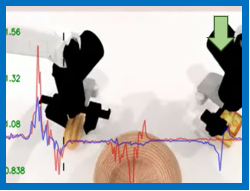}
        \caption{Hesitation resolved; gripper closes}
    \end{subfigure}
    \begin{subfigure}[t]{0.24\textwidth}
        \includegraphics[width=\linewidth]{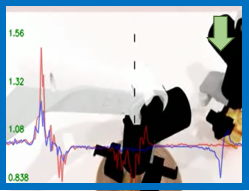}
        \caption{Aiming well}
    \end{subfigure}

    \vspace{0.5em}

    \begin{subfigure}[t]{0.24\textwidth}
        \includegraphics[width=\linewidth]{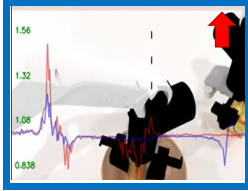}
        \caption{Hesitating over whether to release}
    \end{subfigure}
    \begin{subfigure}[t]{0.24\textwidth}
        \includegraphics[width=\linewidth]{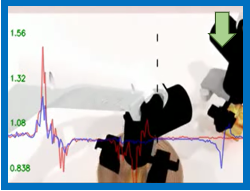}
        \caption{Hesitation resolved; object released}
    \end{subfigure}
    \begin{subfigure}[t]{0.24\textwidth}
        \includegraphics[width=\linewidth]{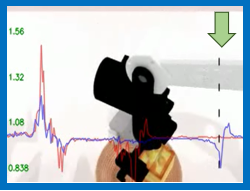}
        \caption{Aiming well}
    \end{subfigure}
    \begin{subfigure}[t]{0.24\textwidth}
        \includegraphics[width=\linewidth]{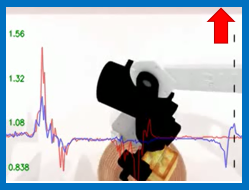}
        \caption{Hesitating over whether to release}
    \end{subfigure}

    \caption{Uncertainty visualization during the \texttt{Place\_Bread\_Basket} task. Variance is low while each arm approaches a clear target (a,d,g), rises during oscillation or hesitation (b,e,h), and decreases after the policy commits to grasping or releasing (c,f). Higher values denote greater uncertainty.}
    \label{fig:uncertainty_timeline}
\end{figure*}

The bread-to-basket sequence in Figure~\ref{fig:uncertainty_timeline} links variance peaks to specific behaviors. Uncertainty remains low during the direct approach in panel (a), rises as the gripper oscillates between its current pose and a feasible grasp pose in panel (b), and falls after the controller commits and closes the gripper in panel (c).

The pattern repeats near the basket: alignment corresponds to low uncertainty in panel (d), hesitation before release produces a spike in panel (e), and completing the release lowers the signal in panel (f). The other arm exhibits a similar transition in panels (g) and (h), suggesting that the peaks are not tied to a particular joint or fixed episode phase.

Together, Figures~\ref{fig:uncertainty_success_failure} and~\ref{fig:uncertainty_timeline} reveal two complementary patterns: failed episodes contain more frequent high-variance intervals, and local peaks coincide with oscillation or delayed commitment. Table~\ref{tab:uncertainty_pretrain_compare} complements these examples with fixed labels, validation-only threshold selection, action-error ranking, and gradient-based warning time.
\end{document}